\documentclass[10pt,journal,compsoc]{IEEEtran}

\usepackage[nocompress]{cite}

\usepackage[pdftex]{graphicx}

\usepackage[caption=false,font=footnotesize,labelfont=sf,textfont=sf]{subfig}
\usepackage{dblfloatfix}

\usepackage{url}

\usepackage{amsfonts}  
\usepackage{amssymb,amsmath}
\usepackage{array,multirow}
\usepackage{blindtext}
\usepackage{bm}
\usepackage{booktabs} 
\usepackage{color}
\usepackage[T1]{fontenc} 
\usepackage[utf8]{inputenc}
\usepackage{mathtools}
\usepackage{microtype}
\usepackage{makecell}

\usepackage{algorithm}
\usepackage{algorithmic}

\newcommand{\comment}[1]{}

\newcommand{\bOne}{{\mathbf{1}}}

\newcommand{\bB}{{\mathbf{B}}}
\newcommand{\bC}{{\mathbf{C}}}

\newcommand{\bP}{{\mathbf{P}}}

\newcommand{\bW}{{\mathbf{W}}}

\newcommand{\bbR}{{\mathbb{R}}}

\newcommand{\cE}{{\mathcal{E}}}
\newcommand{\cL}{{\mathcal{L}}}

\newcommand{\blambda}{\bm{\lambda}}

\usepackage{eso-pic}
\usepackage{xspace}
\makeatletter
\DeclareRobustCommand\onedot{\futurelet\@let@token\@onedot}
\def\@onedot{\ifx\@let@token.\else.\null\fi\xspace}
\def\eg{\emph{e.g}\onedot} 
\def\ie{\emph{i.e}\onedot}

\makeatother

\begin{document}
\title{C2S2: Cost-aware Channel Sparse Selection for Progressive Network Pruning}
\author{Chih-Yao~Chiu, 
        Hwann-Tzong~Chen,~\IEEEmembership{Member,~IEEE,}
        and~Tyng-Luh~Liu,~\IEEEmembership{Member,~IEEE}}
        
\IEEEtitleabstractindextext{
\begin{abstract}
This paper describes a channel-selection approach for simplifying deep neural networks. Specifically, we propose a new type of generic network layer, called pruning layer, to seamlessly augment a given pre-trained model for compression. Each pruning layer, comprising $1 \times 1$ depth-wise kernels, is represented with a dual format: one is real-valued and the other is binary. The former enables a two-phase optimization process of network pruning to operate with an end-to-end differentiable network, and the latter yields the mask information for channel selection. Our method progressively performs the pruning task layer-wise, and achieves channel selection according to a sparsity criterion to favor pruning more channels. We also develop a cost-aware mechanism to prevent the compression from sacrificing the expected network performance. Our results for compressing several benchmark deep networks on image classification and semantic segmentation are comparable to those by state-of-the-art. 
\end{abstract}

}

\IEEEpeerreviewmaketitle 
\maketitle

%
\section{Introduction}
\label{sec:intro}
%

Techniques based on deep neural networks have been very successful for a wide variety of applications in artificial intelligence, but the good performance often comes at a price. A deep-net model typically comprises a huge number of parameters that may not be efficient for run-time computation and also take up large amounts of storage. Such concerns have not been mitigated by that most of the state-of-the-art deep learning methods tend to adopt deeper and complicated architecture to improve the accuracy. Meanwhile, there is also an immediate need for lightweight and efficient versions of deep networks that can be easily ported to and run on embedded systems or mobile devices. We are thus motivated to develop a cost-aware method that can reduce the complexity (\eg, size or FLOPs) of a given network significantly, without compromising its expected performance.

Weight quantization and network pruning are the two most popular ways to alleviate the storage requirement and the computational burden, or to reduce the number of parameters and the number of floating point operations for a deep-net model. This work focuses on devising a network-pruning method that removes redundant weights. A major challenge in pruning a deep network is that most of the weight parameters are correlated and finding the best combination of to-be-pruned weights is simply intractable. There is no general principle to decide directly which weights are negligible to the overall performance. While striving to avoid relying on heuristics, we instead develop a new type of generic network layer that can be plugged into a pre-trained network and then guides itself to systematically and effectively perform network pruning via a learning process.

\begin{figure}[t]
\centering
\subfloat[]{\includegraphics[width=0.45\textwidth]{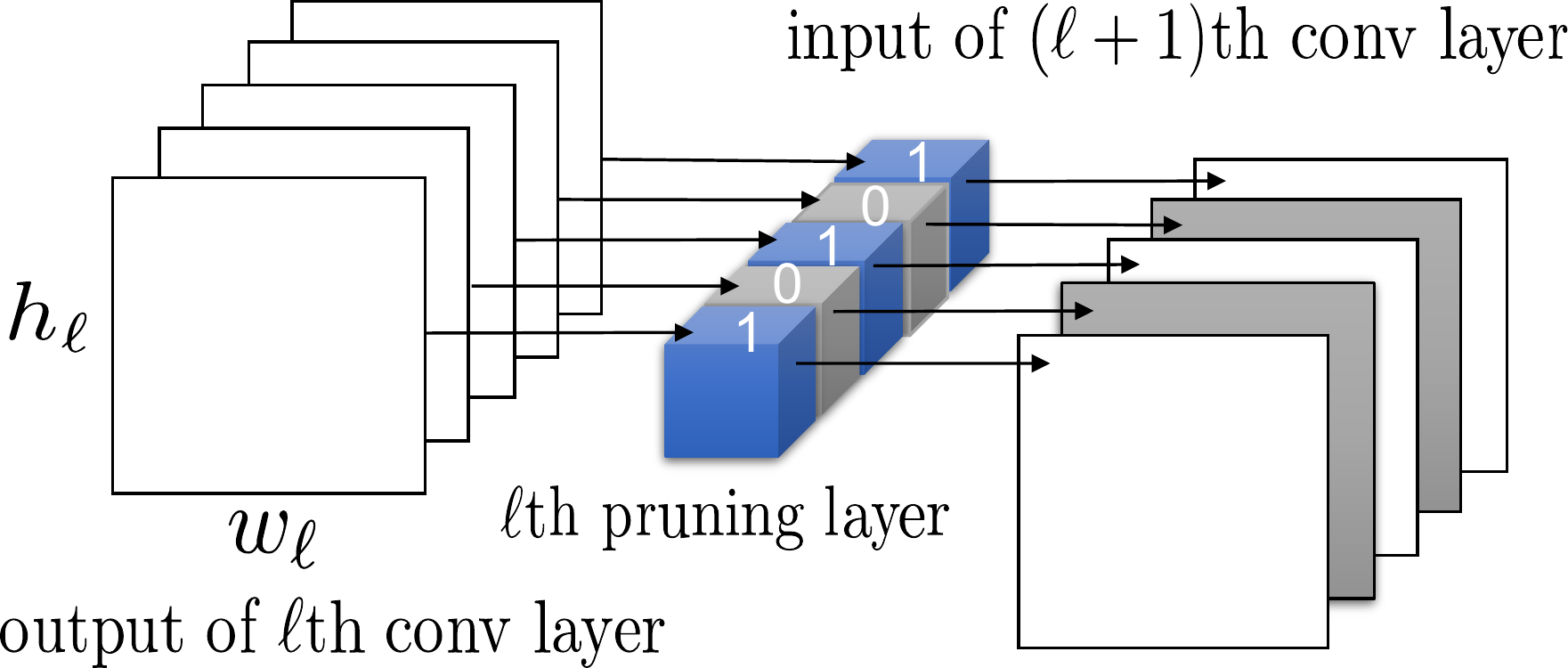}}\\
\subfloat[]{\includegraphics[width=0.20\textwidth]{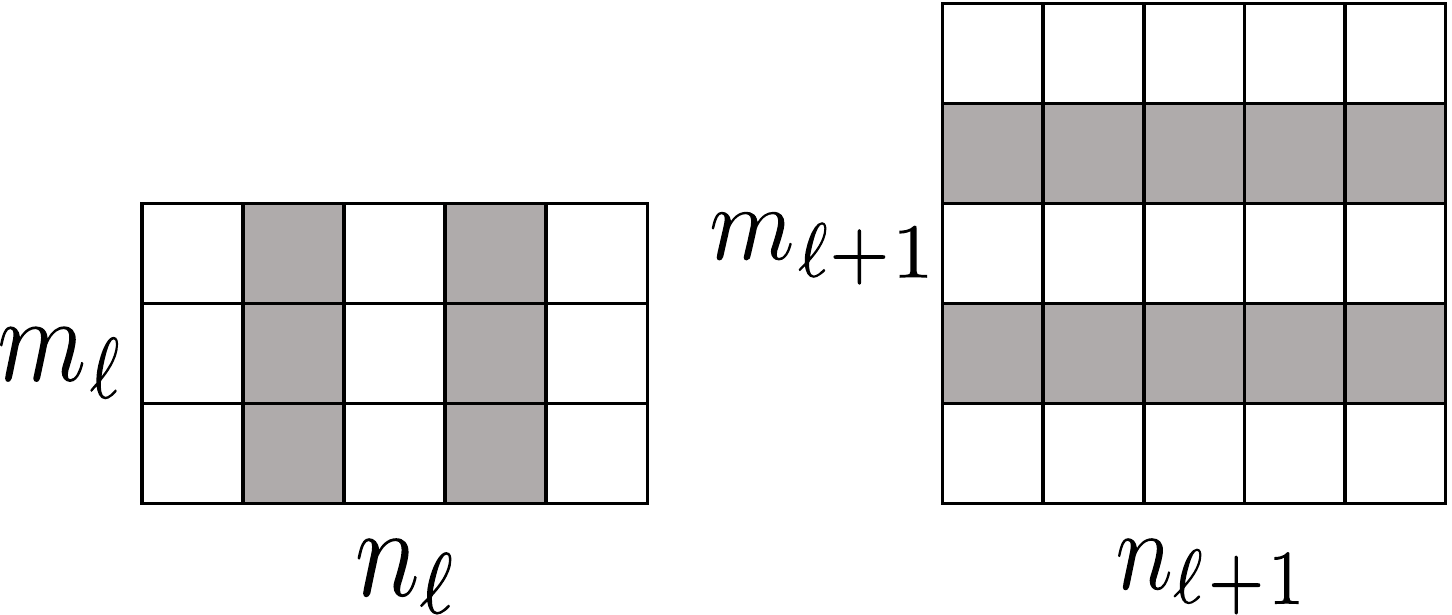}}
\hspace{0.45cm}
\subfloat[]{ \includegraphics[width=0.20\textwidth]{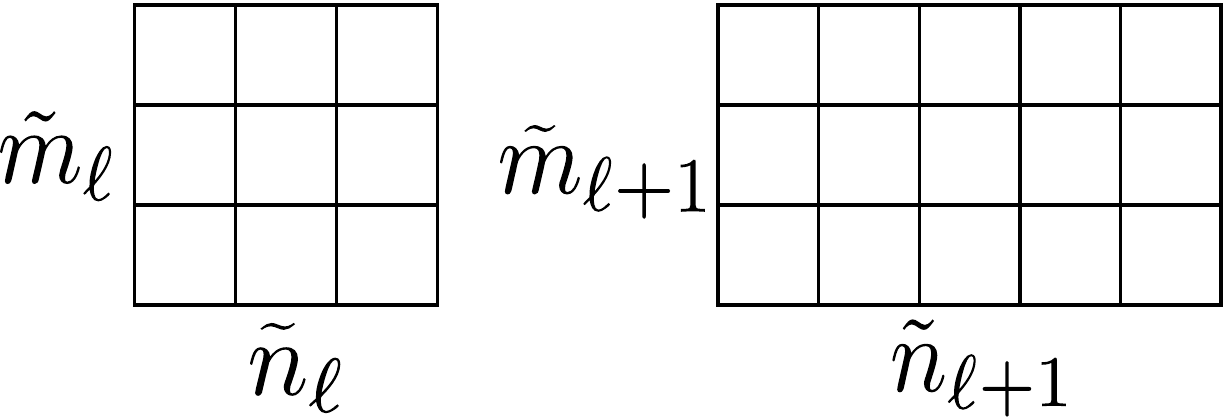}}
\caption{(a) A \emph{pruning layer} is a $1 \times 1$ depth-wise conv layer with binary-valued weights serving as the channel selection mask. (b) The numbers of input and output channels of the $\ell$th layer are $m_{\ell}$ and $n_{\ell}$, respectively, where $n_{\ell} = m_{\ell+1}$. Kernels of $\ell$th layer yielding redundant channels will be removed, while kernels of $(\ell+1)$th layer will be accordingly slimmed. The effect of channel sparse selection is illustrated in (c).}
\label{fig:teaser}
\end{figure}

The proposed new type of layer is called the {\em pruning layer}. Figure~\ref{fig:teaser} illustrates the key idea about how it works. The pruning layer is implemented as a depth-wise convolutional layer with its binary-valued weights serving as the channel selection mask. Filtered by the binary weight values of the pruning layer, those channels that correspond to zero pruning weights will be excluded. As a result, kernels of the previous layer that produce these redundant channels will be removed, while kernels of the next layer will thus be {\em slimmed} according to the values of the pruning mask. To facilitate the pruning, we incorporate a sparsity term into the optimization objective to encourage each pruning layer to remove as many channels as possible. In addition, we also add a bipolar term so that the resulting pruning weights intrinsically tend to have values close to either $0$ or $1$, making it compelling to adopt $0.5$ as the unified threshold for all layer-wise binarizations.

With the proposed pruning layer, we further devise a cost-aware mechanism to avoid excessive pruning of weight parameters. The pruning layer is allowed to overshoot and remove more-than-enough channels. When the overshooting occurs, the training loss tends to be increasing and it may not be feasible to fine-tune the pruned model to yield comparable performance as the original network. The cost-aware pruning mechanism keeps monitoring the changes in the training loss to detect overshooting. When detected, the cost-aware mechanism stops the network-pruning step and  rewinds the training process by restoring some pruned weights back to their previous values. The proposed cost-aware mechanism provides an effective way to compress deep-net models without paying the price of accuracy degradation. We apply our method to different network models, including VGG-Net, ResNet, MobileNet, and FCN. In particular, our method introduces the within-block and between-block pruning steps for ResNet. It is also worth noting that previous approaches to network pruning seldom try to compress deep-net models for tasks other than classification. Our method can efficiently compress FCN for the task of semantic segmentation by removing more than 95\% parameters of the original model while producing comparable results. (See the Appendix.)

%
\section{Related Work}
\label{sec:related}
%

Depending on the design principle, methods of compressing and accelerating deep neural networks could differ significantly. For example, several previous approaches involve learning a more efficient and lightweight network using different architecture or with additional data structures, such as distillation~\cite{HintonVD15}, LCNN~\cite{BagherinezhadRF16}, and MobileNet~\cite{HowardZCKWWAA17}. Our literature survey instead emphasizes those that directly compress existing pre-trained networks. For the ease of discussion, we categorize relevant techniques into two groups: {\em weight quantization} and {\em weight pruning}, and discuss more on the latter as our method for network compression is a pruning-based approach.

\subsection{Weight Quantization}
The technique by~\cite{ZhuHMD16} can achieve nearly $16\times$ compression on several popular networks by quantizing weights into ternary values $\{-W_{\ell}^{n},0,+W_{\ell}^{p}\}$ where $W_{\ell}^{n}$ and $W_{\ell}^{p}$ are learned for each layer. Subsequently, \cite{CaiHSV17} present a quantization based method that uses a half-wave Gaussian quantizer for approximating the ReLU activation non-linearity. \cite{ZhouYGXC17} quantize the full-precision weights into low-precision versions with the values constrained to powers of two or zero. Instead of considering all the weights jointly, they quantize only a portion of the weights of the given layer at a time and regain the accuracy by re-training. These quantization-based methods require specialized software libraries or hardware to accelerate the compressed networks, while our intended approach does not leverage such libraries or hardware for accelerating compressed networks.

\subsection{Weight Pruning}
The concept of weight pruning can be traced back to classic methods on reducing network complexity and improving generalization~\cite{CunDS89,HassibiSW93}, where second-derivative information is used to assess weight importance. Another early work on synaptic pruning \cite{ChechikMR98} suggests to overgrow the synapses first and then prune them with respect to an optimal ``minimal-value" deletion. \cite{LiKDSG16} propose to measure the importance of a filter in each layer by its sum of absolute weights. Their method prunes the least useful filters according to sensitivity analysis for each layer. The drawback is that they have to conduct a series of analyses to determine how many filters can be pruned for each convolutional layer, which is time-consuming especially for very deep networks. Their strategy is to prune each layer independently and evaluate the accuracy. However, the sensitivity of a specific layer might change due to the pruning of other layers. In contrast, our cost-aware channel selection method prunes the network layer by layer and learns how many filters can be pruned without enforcing extra heuristics, which is more convenient and efficient.

In \cite{AnwarHS15,LiKDSG16,ZhouAP16}, the main idea is to evaluate the importance of weights by handcrafted criteria. Our method instead learns the importance of convolutional filters via a generic training process, and can automatically choose the crucial filters to keep without being guided by any handcrafted criteria.

Another line of research aims to learn a compact network by imposing sparsity constraints on the weights of convolutional layers. \cite{AlvarezS16} propose to learn the number of neurons for each convolutional layer by including a group Lasso constraint on the weights. {\em Structured Sparsity Learning} (SSL) by \cite{WenWWCL16} employs group Lasso constraints to multiple groups of weights for channel-wise, shape-wise, and depth-wise pruning. \cite{AghasiANR17} propose the {\em Net-Trim} that uses convex optimization to prune the network layer-wise, and provide theoretical guarantees on how far the pruned model is from the original one. \cite{HeZS17} introduce a coefficient vector for channel selection. They reduce the number of output channels for each layer by imposing an $l_{1}$ norm sparsity constraint on the corresponding coefficient vector. The pruning process is carried out via minimizing the reconstruction error on feature maps and the sparsity constraint. Our method, in comparison, directly minimizes the sparsity constraint and loss between the prediction and label, yielding a much simpler optimization problem. \cite{LuoWL17} develop the so-called ThiNet model for filter level pruning. It respects the network structure, and also performs filter pruning layer-by-layer. The pruning strategy is an iterative scheme that decides the pruning of layer $i$ based on the statistics of layer $i+1$ and optimizes a reconstruction error caused by the filter selection. However, in our method, the layer-wise pruning order can be arbitrary, and the pruning process is driven by preserving the network performance.

Network slimming by \cite{LiuLSHYZ17} shares a similar idea with our approach in that the common goal is to achieve channel-level sparsity and the insignificant channels are identified via a training process. They impose $l_1$ regularization on the scaling factors in batch normalization layers to achieve network compression. The key differences between their approach and ours are as follows:
\emph{i}) We introduce the cost-aware mechanism for recovering from over-pruning. The proposed {\em Cost-aware Channel Sparse Selection} (C2S2) method can automatically maintain the accuracy of the network during the process of progressive pruning. That is, it learns to generate the final compressed network in one pass by taking account of the trade-off between compression ratio and performance degradation. \emph{ii}) Our approach includes a bipolar term in the loss function to enforce the convergence of pruning weights toward either $0$ or $1$ for easy binarization using a fixed threshold ($\tau = 0.5$). \emph{iii}) The proposed C2S2 carries out channel pruning layer by layer so that only kernels of a same layer are considered simultaneously. The strategy is more advantageous over global pruning for keeping the network architecture intact. In contrast, the method of \cite{LiuLSHYZ17} prunes channels by specifying a global percentile of sorted scaling factors across all layers. The pruning process is thus driven by the specified compression rate. It requires several full-fledged runs to decide the most suitable compression ratio that the resulting compressed network  can still be fine-tuned to yield comparable performance to the original network. Our method instead performs channel selection with a cost-aware mechanism to layer-wise pursue most effective pruning with the constraint of maintaining the network performance. More discussions can be found in the {\bf Experiments}.

%
\section{Progressive Network Pruning}
\label{sec:progressive}
%

\begin{figure*}[t]
\centering
\subfloat[Pruning with augmented network] 
{\includegraphics[width=0.45\textwidth]{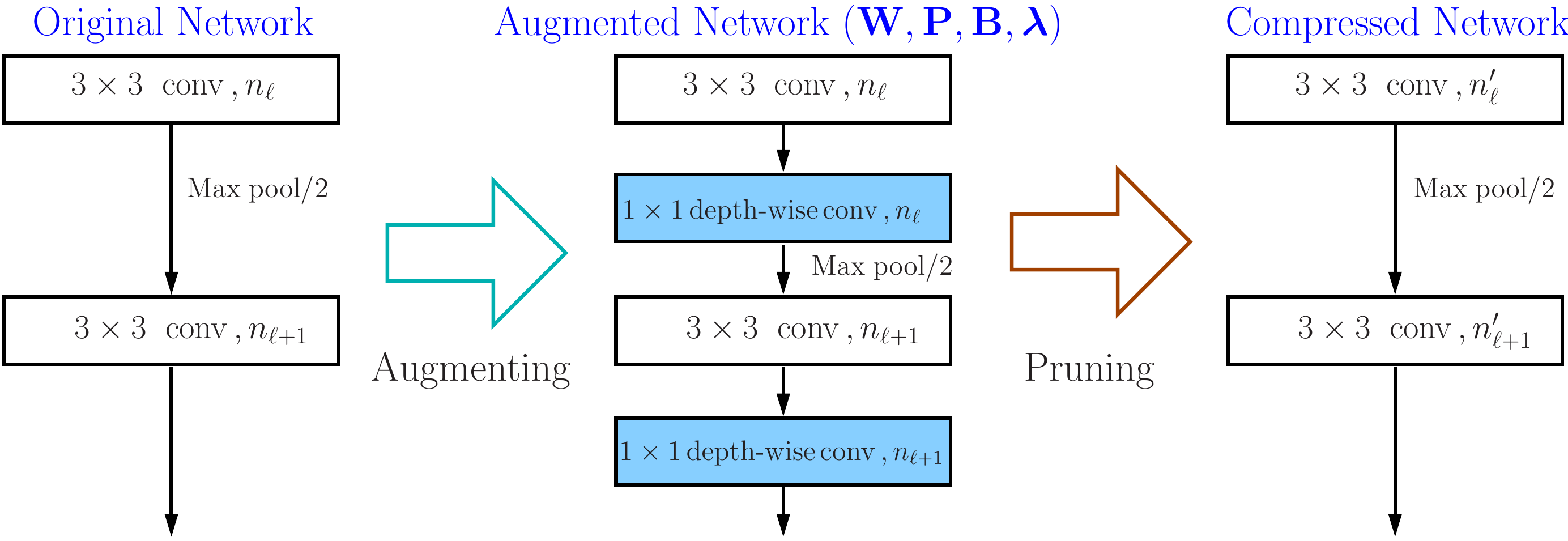}}
\hspace{0.25cm}
\subfloat[Within-block pruning] {\includegraphics[width=0.25\textwidth]{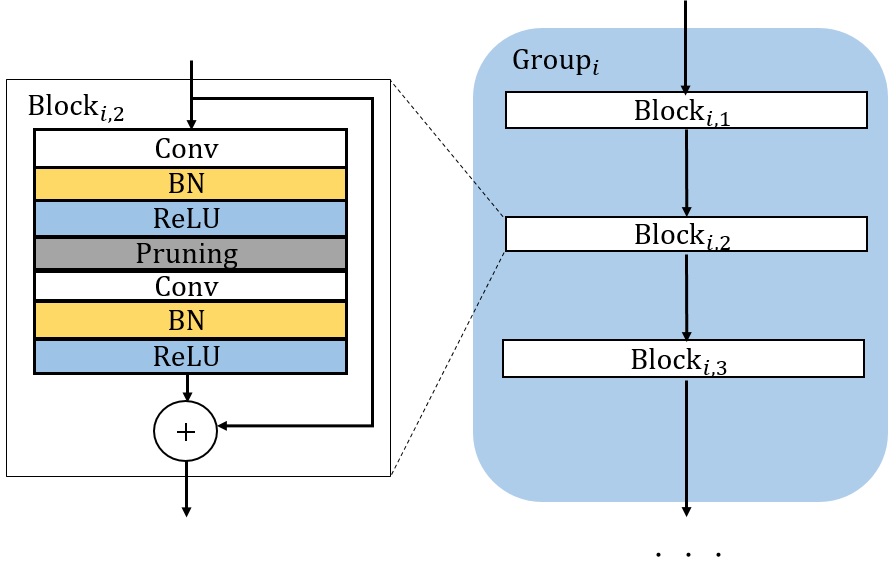}}
\hspace{0.05cm}
\subfloat[Between-block pruning] {\includegraphics[width=0.25\textwidth]{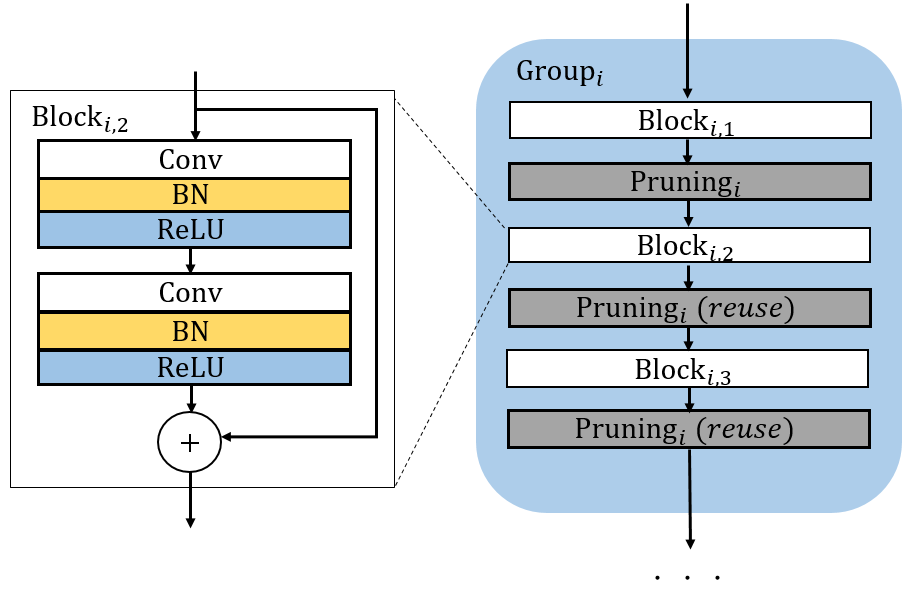}}
\caption{(a) Given a ConvNet, an augmented network is constructed by inserting a pruning layer next to each conv layer. After performing channel sparse selection on the augmented network, we layer-wise unplug those inserted layers and remove the masked channels. (b)-(c) C2S2 for residual networks: Denote the $j$th residual block in the $i$th residual group as Block$_{i,j}$.  (b) Within-block pruning. (c) Between-block pruning: Insert a pruning layer after each residual block and make the pruning layers of the same group share the same weights. After pruning, unplug all pruning layers and remove redundant channels for each layer.}
\label{fig:new}
\end{figure*}

We explore the layer-wise architecture of deep network for network compression. The strategy, as we will describe below, could lead to not only a general framework but also an effective approach. At the core of our method is a new type of generic network layer called {\it pruning layer} that plays the crucial role of selecting and masking channels during the pruning process. The pruning layer is considered generic in that it is a flexible {\em depth-wise} module and can accommodate almost all types of neural networks. After inserting the pruning layers to a neural network, the function of each pruning layer is to select {\em informative channels} from its inputs, \ie, the outputs from its preceding layer. Hence the pruned channels can be either feature maps or simply neurons, depending on whether a pruning layer is positioned right after a convolutional layer or a fully-connected layer. However, in this study we restrict the use of pruning layers to expunging redundant feature maps from convolutional layers.

Given a pre-trained ConvNet with $L$ convolutional layers, we use  $\bW = \{W_\ell\}_{\ell=1}^L$  to denote the set of relevant network parameters. The layer-wise outputs are expressed by $\bC = \{C_\ell\}_{\ell=1}^L$, where tensor $C_\ell$ includes the feature maps yielded by the $\ell$th convolutional layer. We augment the model with pruning layers, each of which follows right next to a convolutional layer. After performing channel sparse selection on the augmented network, we unplug the pruning layers and remove the masked channels to obtain a compressed network with $\widetilde{\bC} = \{\widetilde{C}_\ell\}_{\ell=1}^L$ that contains a reduced number of channels for each convolutional layer.  We illustrate the overall pruning process in Figure~~\ref{fig:new}a.

Specifically, each pruning layer comprises an array of $1 \times 1$ depth-wise kernels, where the total number is specified by that of its input channels. We denote the real-valued weight tensor of the pruning layer following the $\ell$th convolutional layer as $P_\ell$. The goal of our network pruning is to optimize $\bP = \{P_\ell\}_{\ell=1}^L$ so that the real-valued weights can be automatically binarized with respect to a {\em unified} threshold $\tau$ to yield binary masks
$\bB = \{B_\ell\}_{\ell=1}^L$. With the above notations, the corresponding augmented network can be represented by a quadruplet $(\bW, \bP, \bB, \blambda)$, where $\blambda=\{\lambda_1, \lambda_2\}$ are the weight parameters in the objective function (\ref{eqn:loss}) for the layer-wise pruning. (By a slight abuse of notation, $\bW$ in the quadruplet form indeed means to represent all trainable weights of the given ConvNet.)
We can write the process of channel selection (network pruning) as
\begin{equation}
\widetilde{\bC} = \bC \odot \bB = \{C_\ell \odot B_\ell\}_{\ell =1}^L \,,
\label{eqn:compress}
\end{equation}
where $\odot$ symbolizes the channel-wise product between $C_\ell$ and $B_\ell$. From (\ref{eqn:compress}), we see that $B_\ell$ with its values from $\{0, 1\}$ functions as the layer-wise binary channel selection mask.

\subsection{Channel Sparse Selection}

With the augmented network, the proposed channel selection algorithm starts with a training process that aims to learn the binary channel selection masks $\bB$. To this end, we consider a progressive scheme that carries out the pruning task in a specific layer-wise order, which we have considered three implementations: {\em forward} , {\em backward}, and {\em interlaced} orderings. Since the channel selection process is analogous for each pruning layer, we omit the layer subscript $\ell$ in the following analysis for the sake of simplifying the notations.

Let the data ground truth for the underlying task that the given ConvNet aims to accomplish be $Y$. To perform channel selection for a specific pruning layer, our algorithm operates in two phases. In the first phase, the goal is to optimize the real-valued weights $P$ of the pruning layer, while all other network parameters $\bW = \{W_\ell\}_{\ell=1}^L$ are fixed. Then, in the second phase, while fixing $P$, our algorithm optimizes all the {\em remaining} convolutional parameters according to the selection mask $B$, which is derived from $P$ with respect to the threshold $\tau = 0.5$. We detail the two phases of performing network pruning below.

\subsubsection{Phase 1}
With $\bW$ fixed, we impose a sparsity constraint and a bipolar prior for optimizing $P$, the real-valued weights of the pruning layer.  Specifically, the objective function for layer-wise pruning can be written as
\begin{equation}
	\cL (Y,Y_1)+\lambda_{1}\cdot \|P \|_{1} + \lambda_{2} \cdot \|P \odot (\bOne - P)\|_{1} \,,
    \label{eqn:loss}
\end{equation}
where $\bOne$ is the unit tensor with the same size of $P$, and $\cL$ is a function (to be specified later) measuring the {\em difference} between ground-truth labels $Y$ and phase-1 predictions $Y_1$. The $\| P \|_{1} $ term is to favor the sparsity of the pruning layer's weights, \ie, to  prefer pruning more channels. The second $l_1$-norm term encourages the $1 \times 1$ kernels of $P$ converging to $0$ or $1$, and thus computing the corresponding binary mask $B$ is not sensitive to the choice of threshold $\tau$. So, in the first phase, we do forward propagation through the augmented network $(\bW, \bP, \blambda)$, and compute the total loss according to ~(\ref{eqn:loss}). Notice that $\bB$ is intentionally left out of the quadruplet notation since binary masks are not considered in the first phase. As a result, we have an end-to-end differentiable neural network. In back-propagation, only $P$ are updated with respect to the gradient feedback.

It is convenient to reshape the tensor $P \in \bbR^{1\times1\times n\times1}$ into a 1-D vector $P \in \bbR^{n}$, where $n$ is the number of output channels of the convolutional layer that the pruning layer is plugged right after. (See Figure~\ref{fig:teaser} for illustration.) We then use $P(i)$ to denote the $i$th element of $P$. In optimizing $P$, although we do not enforce to clamp the value of each $P(i)$ to fall between 0 and 1,  the value can still be considered as the response for supporting the selection of convolutional output channel $i$. It follows that each element $B(i)$ of binary mask $B$ signals the channel selection for the $i$th output channel of the preceding convolutional layer.

\subsubsection{Phase 2}
With $\bP$ fixed, the second phase of our algorithm begins by specifying the binary mask $B$. Specifically, we compare $P(i)$ with threshold $\tau=0.5$ to decide $B(i)$:
\begin{equation}
B(i)=\left\{ \begin{array}{cl} 1\,, & \mbox{if \, $P(i)>\tau$ \,;} \\ 0\,, & \mbox{otherwise.} \end{array}\right.
\label{eqn:mask}
\end{equation}

In the second phase, we do forward propagation through the ConvNet architecture filtered by $B$ and obtain the predictions $Y_2$. As $P$ is fixed in phase two, the total loss in (\ref{eqn:loss}) is reduced to $\cL(Y, Y_2)$. During back-propagation, we fix the weights of each pruning layer and update the weights of each convolutional layer and each fully connected layer. In our implementation, phase one is carried out every 10 training steps, while phase two is done for each training step.

\comment{
\begin{figure*}[t!]
	\centering
    \subfloat[] { \includegraphics[width=0.35\textwidth]{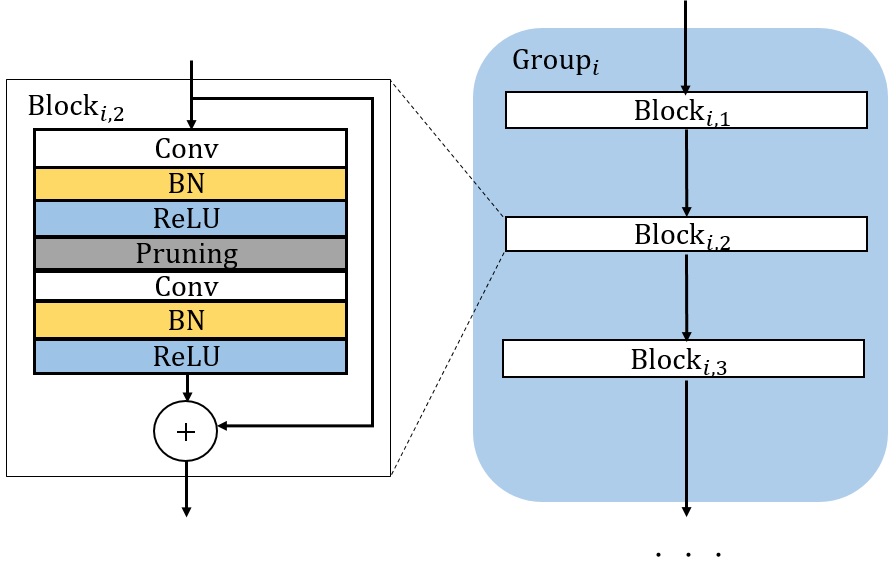} }
    \hspace{0.75cm}
    \subfloat[] { \includegraphics[width=0.35\textwidth]{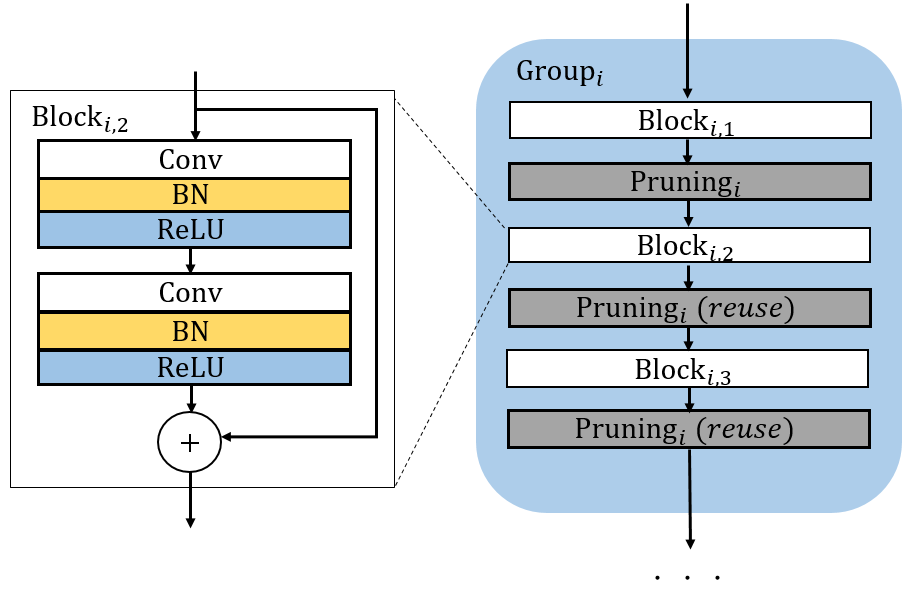} }
    \caption{C2S2 for residual networks. Denote the $j$th residual block in the $i$th residual group as Block$_{i,j}$.  (a) Within-block pruning. (b) Between-block pruning: We insert a pruning layer after each residual block and make the pruning layers of the same group share the same weights. After pruning, we unplug all pruning layers and remove redundant channels for each layer. }
    \label{fig:figure3}
\end{figure*}
}

\subsubsection{Channel Selection and Kernel Pruning/Slimming}
The relation between channel selection and kernel pruning/slimming is illustrated in Figure~\ref{fig:teaser}. If a particular output channel is excluded by the binary mask, say, $B_\ell$, the respective kernel of the $\ell$th convolutional layer that produces this redundant channel will be pruned, while the kernels in the $(\ell+1)$th layer will be slimmed by one input channel. In other words, the kernels of $\ell$th layer that produce redundant channels will be removed, while kernels of $(\ell+1)$th layer will be slimmed by the number of removed channels.

\subsubsection{C2S2 for Residual Networks}
Our method can also be used to compress residual networks~\cite{HeZRS16}. Since residual blocks with the same number of output channels are distributed consecutively in a residual network, we thus divide them into (consecutive) residual groups with respect to the number of their output channels. Now consider an arbitrary group. If we prune any output channels of its last residual block, then, due to the design of identity shortcut, we also need to prune the corresponding input channels of that block to maintain the channel consistency. This will cause a chain effect such that the corresponding output channels of each residual block in the particular group must be removed accordingly. It implies that the pruning layers that we add right after each residual block of a common group must be made to yield the same binary mask. Further, to prevent the chain effect from propagating to the preceding groups, we implement the shortcut of the first residual block in each group as a  $1\times 1$ convolutional layer. We thus propose a two-step pruning for each group. In the first step, as in Figure~\ref{fig:new}b, we insert a pruning layer after the first convolutional layer of each block and carry out within-block channel pruning. In the second step, we perform between-block pruning. Since all residual blocks within a group have the same number of output channels, they can be pruned by using a shared pruning layer. (See Figure~\ref{fig:new}c.) Likewise, we can obtain the final compressed residual network. The flexibility for performing between-block pruning is one of the key advantages of the proposed pruning layer, while other approaches, \eg, \cite{LiuLSHYZ17,LuoWL17}, work solely on the original network architecture and are prevented from doing so due to the short-cut connections.

\begin{figure*}[t]
  \centering
  \includegraphics[width=0.27\linewidth,height=0.22\linewidth]{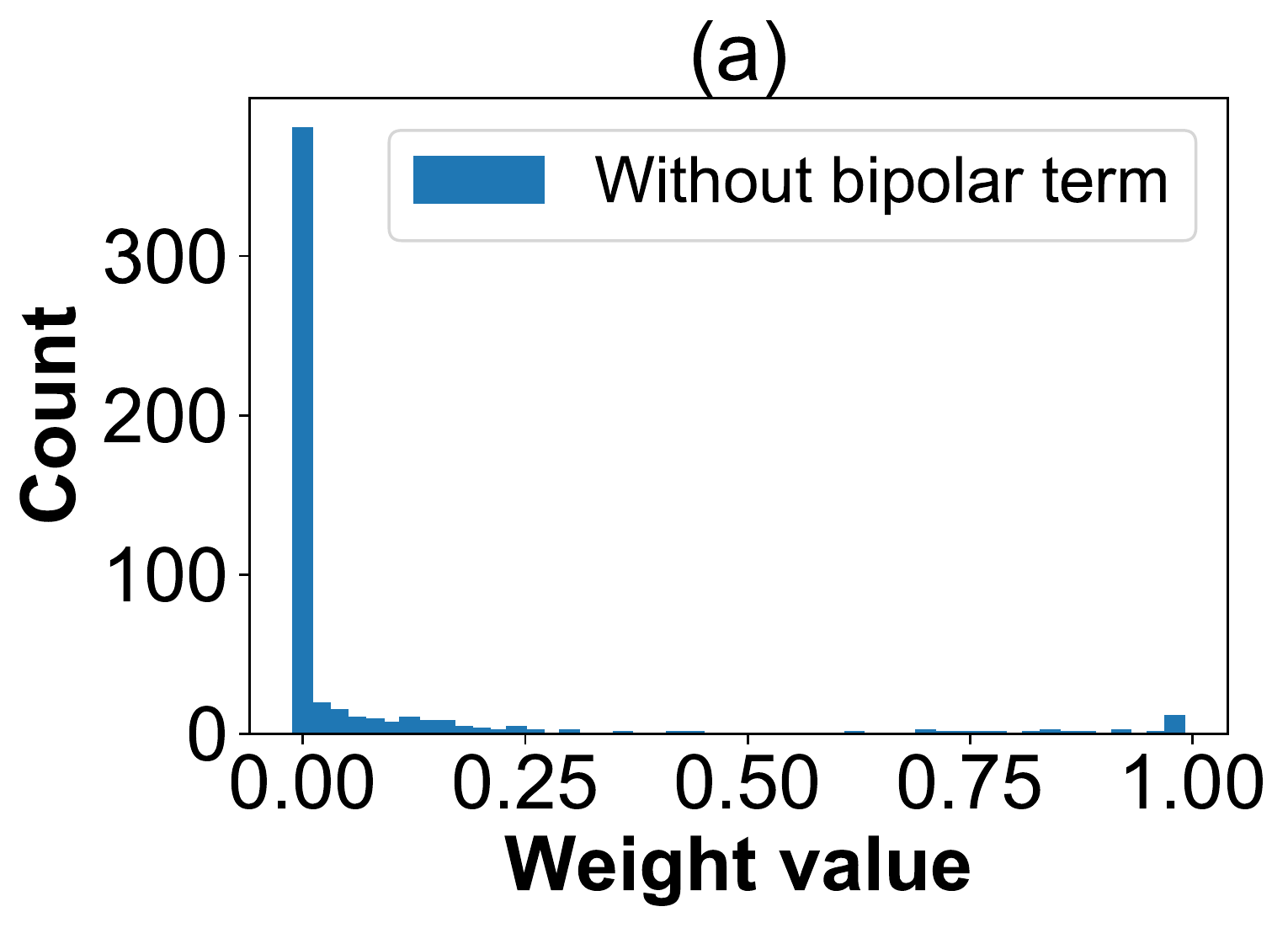}
  \includegraphics[width=0.27\linewidth,height=0.22\linewidth]{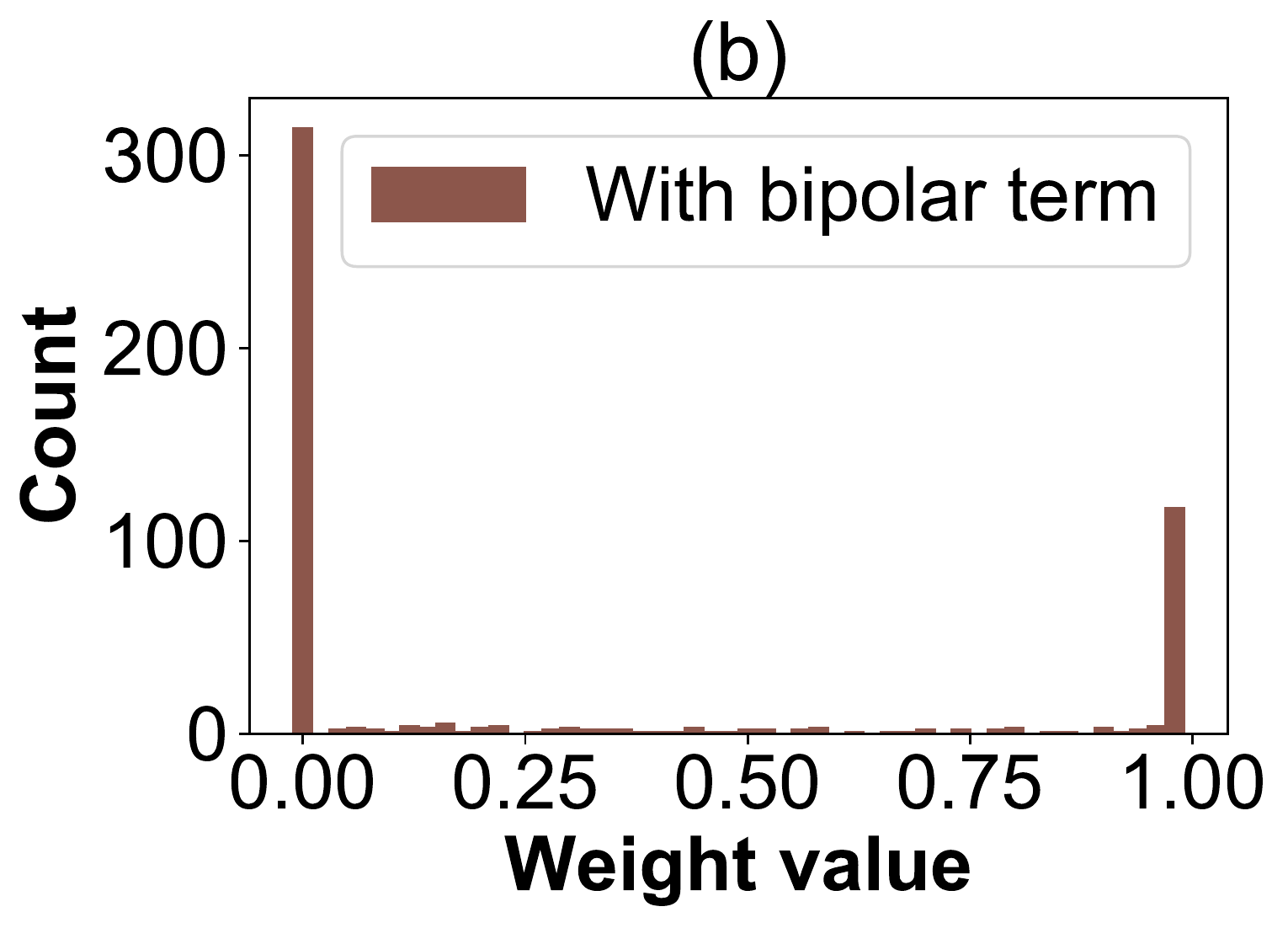}
  {\setbox1=\hbox{\includegraphics[width=0.40\linewidth,height=0.22\linewidth]{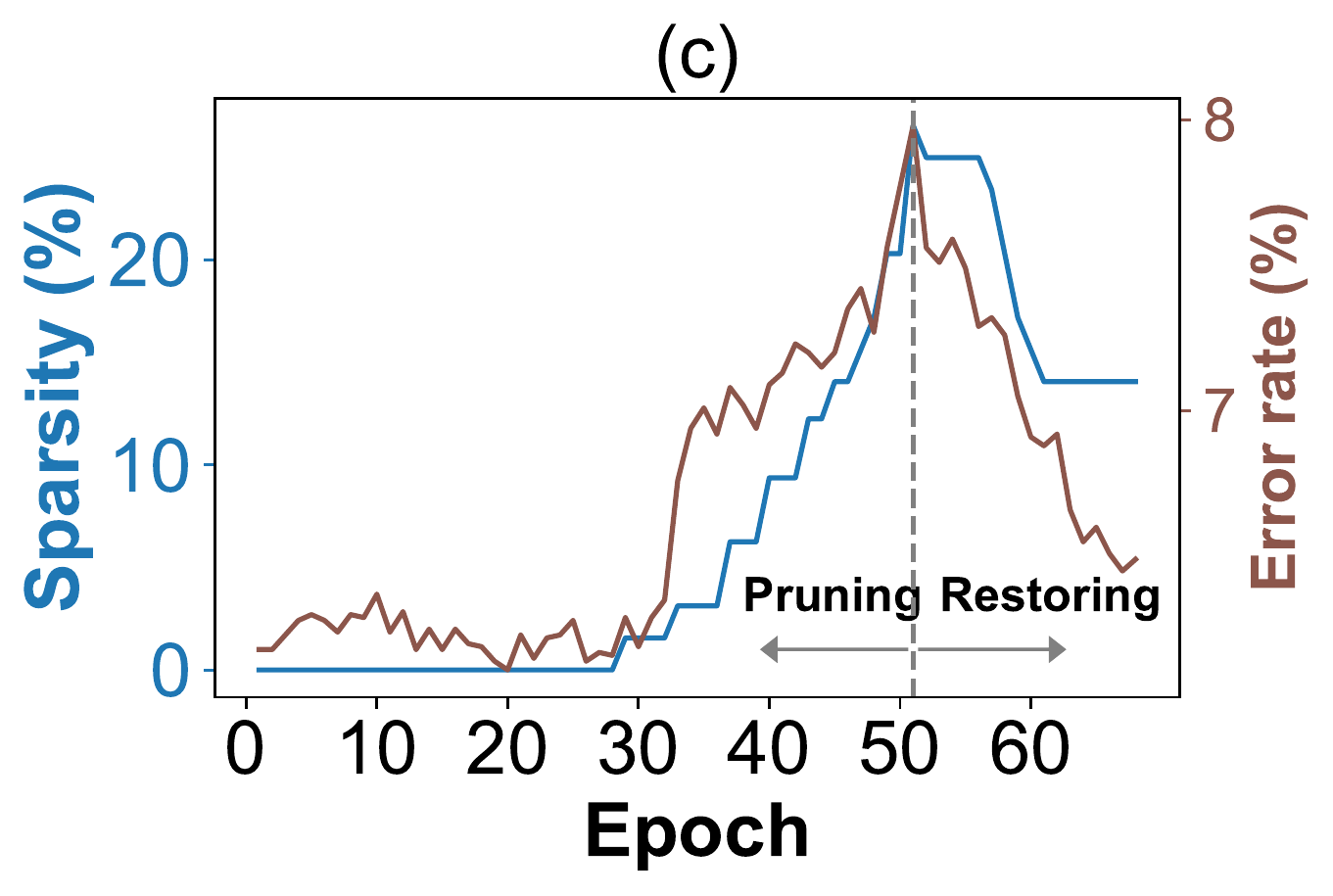}}
\includegraphics[width=0.40\textwidth,height=0.22\linewidth]{fig/acc_spa.pdf}\llap{\makebox[0.83\wd1][l]{\raisebox{2.4cm}{\includegraphics[width=0.18\linewidth]{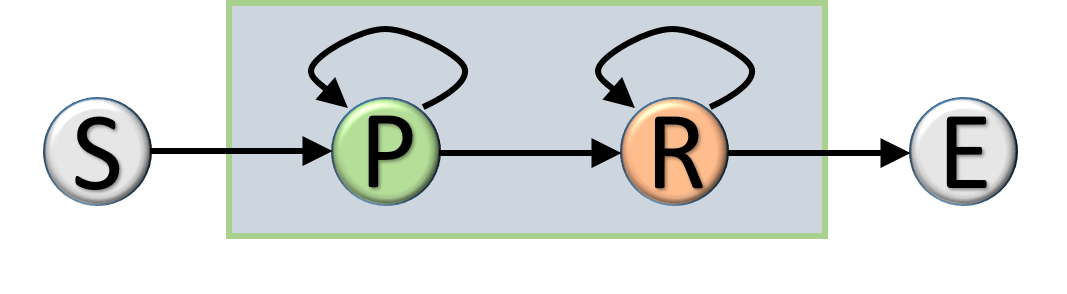}}}}}
  \caption{(a)-(b) The impact of bipolar prior on the weights of  pruning layers: We show the pruning weight distribution for VGG-Net Conv16 when training on CIFAR-10 (a) with only the sparse constraint $\lambda_{1} = 0.002$, or (b) with both the sparsity and bipolar constraints $\lambda_{1} = \lambda_{2} = 0.002$. (c) Upper-left is the state transition diagram, illustrating the C2S2 mechanism, where S, P, R, E correspond to the {\tt Start}, {\tt Pruning}, {\tt Restoring}, and {\tt End} state respectively. With cost-aware, the pruning process detects over-pruning at epoch 51 to trigger the state transition from {\tt Pruning} to {\tt Restoring} for retaining network performance.  } 
   \label{fig:combo}
\end{figure*}


\section{Cost-aware Mechanism}
\label{sec:cost-aware}


Previous work, \eg, \cite{LiKDSG16} mainly estimates how many insignificant parameters can be pruned from each network layer by conducting a series of sensitivity analysis. In our approach, we attempt to automatically determine the proper sparsity for each convolutional layer via a {\em cost-aware} mechanism. As a result, our method can compress a given ConvNet in a much simpler and efficient way. 

The cost-aware mechanism can be thought of as a simple case of two-state Markov decision process (MDP). As illustrated in Figure~\ref{fig:combo}c, the process of channel selection has two (operation) states, namely, {\tt Pruning} and {\tt Restoring}, whereas the action to take at the former is to decide how many channels to prune, and at the latter is to estimate how many pruned channels should be restored. We now describe a greedy and heuristic approach to the process. To begin with, we run all the training examples with the given pre-trained ConvNet and obtain the average error rate $\cE_{base}$. Then, with the augmented network, we maintain an incrementally updated error rate, denoted as $\cE_{ema}$, which is an {\em exponential moving average} of the training error rate .

At the beginning of pruning a specific convolutional layer, we trigger the state transition to move from {\tt Start} to {\tt Pruning}. Meanwhile, we set a scaling coefficient $c_{p}$ to form a tolerable upper bound $c_{p} \cdot \cE_{base}$ that is used to prevent $\cE_{ema}$ from increasing too much due to pruning too many channels. Specifically speaking, the process keeps returning to {\tt Pruning} and the algorithm continues to prune redundant channels until $\cE_{ema}$ exceeds $c_{p} \cdot \cE_{base}$. It signals the detection of excessive pruning and causes the process to make the state transition to {\tt Restoring}.

Once the process is at {\tt Restoring} state, our method automatically restores the pruned channels by changing the sign of $\lambda_1$. Particularly, we set a proper coefficient $c_{r}$ such that the algorithm continues to restore the pruned channels until $\cE_{ema}$ is smaller than $c_{r}\cdot \cE_{base}$ and then moves to the {\tt End} state to conclude the channel selection for the convolutional layer. Detailed steps of the C2S2 algorithm for network compression are described in Algorithm 1 in the appendix.


\section{Experiments}
\label{sec:results}


We evaluate our method on several popular ConvNets, including VGG-Net~\cite{SimonyanZ14a}, ResNet~\cite{HeZRS16}, MobileNet ~\cite{HowardZCKWWAA17} and FCN~\cite{Long2014}. To decide the order of pruning, we carry out a pilot experiment on ResNet-20 with a model pre-trained on CIFAR-10 \cite{CIFAR10}. Next, we discuss results and comparisons for experiments on CIFAR-10, CIFAR-100, and ImageNet. Besides the various classification tasks, our method is also applied to the compression of ConvNets for semantic segmentation. 

\subsection{More on C2S2}

We first provide some insightful observations to manifest the advantages of the two key components in C2S2, namely, the bipolar prior in (\ref{eqn:loss}) and the cost-aware mechanism. Figure~\ref{fig:combo}b and Figure~\ref{fig:combo}a show the distributions of pruning weights derived with or without using the bipolar term. It can then be inferred that the weight distributions by C2S2 entail a clear-cut fixed threshold at $0.5$, while those from using only the sparsity constraint do not display such a favorable property. We next highlight the usefulness of pruning with the cost-aware mechanism that enables our method to perform channel selection under the constraint of maintaining network performance. As shown in Figure~\ref{fig:combo}c, the proposed C2S2 can automatically detect the occurrence of over-pruning and makes the state transition from {\tt Pruning} to {\tt Restoring} so that the network performance after sparse channel selection for each layer is reasonably retained. 

The two foregoing nice features enable our method to decompose the overall problem of channel selection into a sequence of layer-wise subproblems. At each layer, the task is to generate a binary mask based on the fixed threshold so that the resulting channel selection can be effectively achieved without degrading the expected accuracy.

\begin{figure}[!t]
\centering
\includegraphics[width=0.485\linewidth]{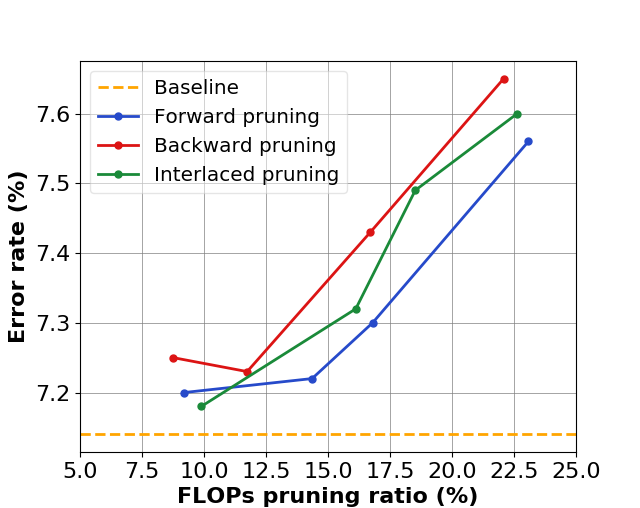}
\includegraphics[width=0.485\linewidth]{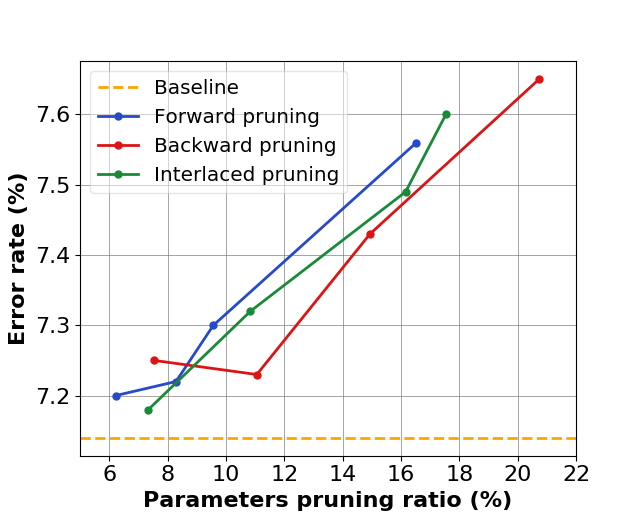}
\caption{ Pruning results of ResNet-20 on CIFAR-10 with different pruning orders. We decide to perform pruning in forward direction since it attains the best accuracy-FLOPs trade-off. Note that the order of interlaced pruning is the first, the last, the second, the second last layer and so on. } 
\label{fig:402}
\end{figure}

Since we prune convolutional layers progressively, the effectiveness may be affected by the pruning order. We thus perform a pilot experiment on ResNet-20 pre-trained on CIFAR-10. The result is reported in Figure~\ref{fig:402}. The channel selection is done in three different orderings, namely, forward, backward, and interlaced.  According to the results of this pilot experiment, we have found that forward pruning yields higher pruning ratios for low-level layers than high-level layers, which enables forward pruning to achieve higher FLOPs reductions than backward and interlaced pruning with roughly the same error rate. In contrast, backward pruning has higher pruning ratios for high-level layers than low-level layers, which results in higher parameters reductions than forward and interlaced pruning given roughly the same error rate. We choose to perform network pruning in forward direction for all our experiments since it attains the best accuracy-FLOPs trade-off (unless otherwise specified).

The overall pruning ratio and accuracy loss are mainly affected by the hyper-parameter $c_{r}$, enforcing the training error rate of a pruned model to be under a certain level. In contrast, the impact of  $\lambda_{1}$ and $\lambda_{2}$ on overall pruning ratio is negligible unless we use $\lambda_{1}$ and $\lambda_{2}$ of different orders. Note that $\lambda_{1}$ and $\lambda_{2}$ are fixed at 0.002 in all experiments.

\subsection{VGG-Net on CIFAR-10}

Our VGG-Net is a variant of VGG-16~\cite{SimonyanZ14a}. We adopt the network architecture described in \cite{LiuLSHYZ17} and implement a VGG-Net that contains 16 convolutional layers and 1 fully connected layer. The detailed architecture of the VGG-Net can be found in Appendix. With the VGG-Net, the function $\cL(Y,Y_1)$ in (\ref{eqn:loss}) is to measure the cross entropy between label $Y$ and prediction $Y_{1}$.

\begin{figure}[!t]
\centering
\includegraphics[width=0.485\linewidth]{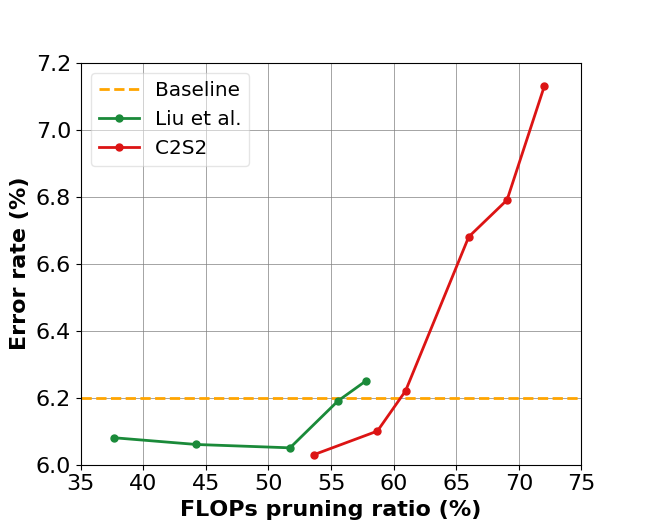}
\includegraphics[width=0.485\linewidth]{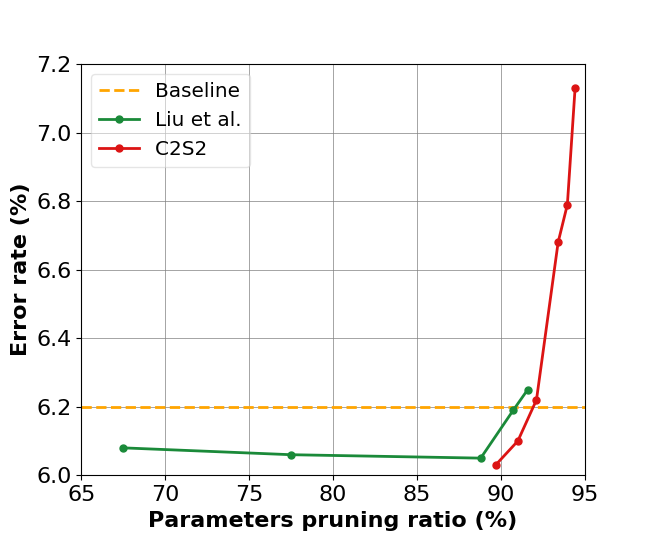}
\caption{Pruning results of VGG-Net on CIFAR-10. We find that \cite{LiuLSHYZ17} will prune all output channels of the 12th convolutional layer when the pre-defined pruning ratio is set too high. In contrast, our method does not suffer form such a problem owing to the cost-aware mechanism.} 
\label{fig:403}
\end{figure}

From the results shown in Figure~\ref{fig:403}, we observe that C2S2 has the noticeable advantage over \cite{LiuLSHYZ17} in achieving pruning without degrading accuracy. In our experiment, C2S2 can easily perform pruning without degrading test accuracy by setting the hyper-parameter $c_{r}$ slightly larger than 1.0.
As shown in the Appendix, C2S2 yields $91\%$ reduction of model parameters with improved accuracy by setting $c_{r}$ as 1.2. In contrast, the technique of \cite{LiuLSHYZ17} requires a pre-defined pruning ratio to do network pruning. However, there is no rule of thumb to choose a proper pruning ratio. In practice, it requires many time-consuming tries to decide the proper value of pruning ratio.     

The other disadvantage from the model of \cite{LiuLSHYZ17} is about retaining the original network architecture. In the experiment, it reaches the highest pruning ratio of parameters at $91.6\%$. When the pruning process continues to carry out and thus increases the pruning ratio, their technique would prune all output channels of the 12th convolutional layer and cause the model to break down. In contrast, C2S2 does not suffer from this problem because the cost-aware mechanism can prevent the pruning process from over-pruning.

Also detailed in the provided Appendix, the proposed method can obtain $92.1\%$ reduction of model parameters and $61.0\%$ reduction of FLOPs for VGG-Net without degrading test accuracy by performing channel sparse selection in forward direction. Our results of reduction in model parameters and FLOPs are superior to those of \cite{LiuLSHYZ17} under the same baseline accuracy and network settings.

\subsection{ResNet-20 on CIFAR-10}

The ResNet-20 architecture adopted in the comparison is the same as that described in \cite{LiuLSHYZ17}. The structure of residual block is plotted in Figure ~\ref{fig:new}b. It is worth emphasizing that we use a $1\times 1$ convolutional layer to do projection mapping in the first residual block of each group. With the projection mapping, we can easily handle the dimension mismatch caused by channel pruning via reducing the output channels of $1 \times 1$ convolutional layer. $\cL(Y,Y_1)$ in (\ref{eqn:loss}) is to measure the cross entropy between labels and predictions. In the first step of C2S2 for residual network, we carry out within-block channel pruning. In the second step, we perform between-block pruning. In contrast, \cite{LiuLSHYZ17} performs only within-block channel pruning on ResNet. 

\begin{figure}[!t]
\centering
\includegraphics[width=0.485\linewidth]{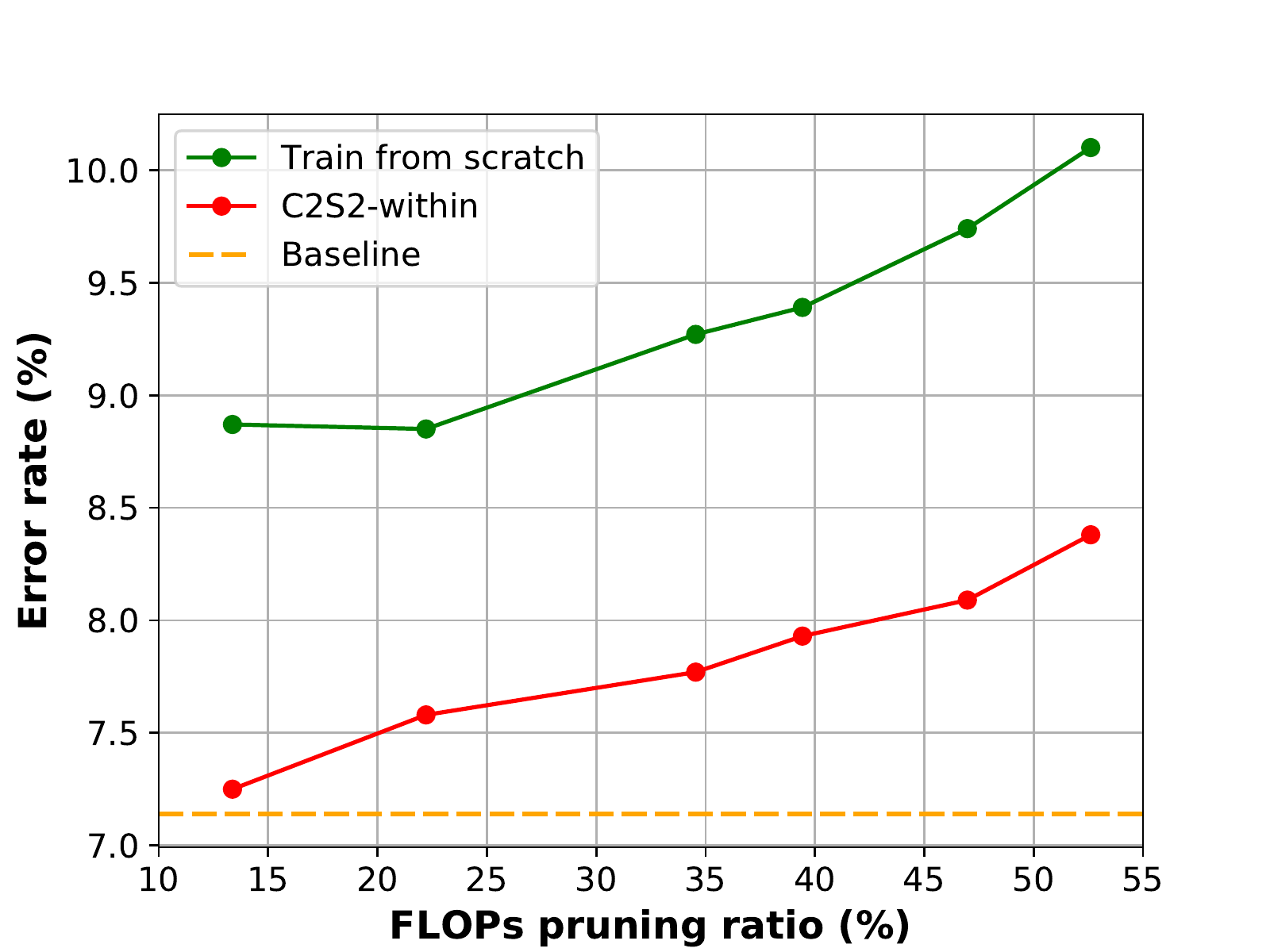}
\caption{ The training results of ResNet-20 on CIFAR-10. We find that directly fine-tuning the pruned models gives better performance than re-training from randomly re-initialized weights. In other words, our pruning process is not just an architecture search method. It preserves the important weights that are beneficial to the performance. } 
\label{fig:retrain_vs_c2s2}
\end{figure}

\begin{figure}[!t]
\centering
\includegraphics[width=0.485\linewidth, height=0.395\linewidth]{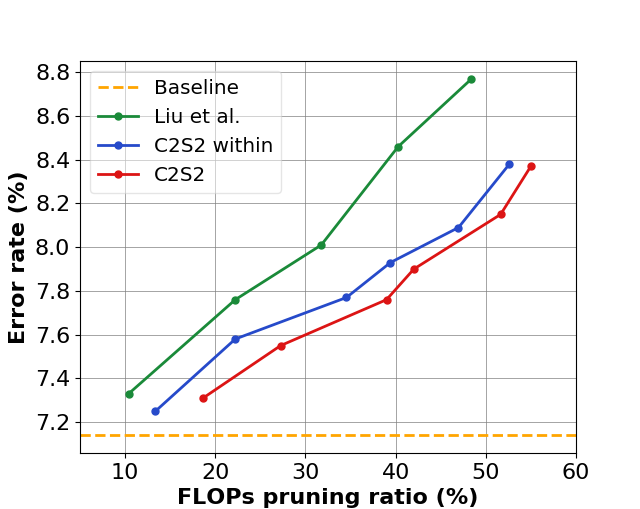}
\includegraphics[width=0.485\linewidth]{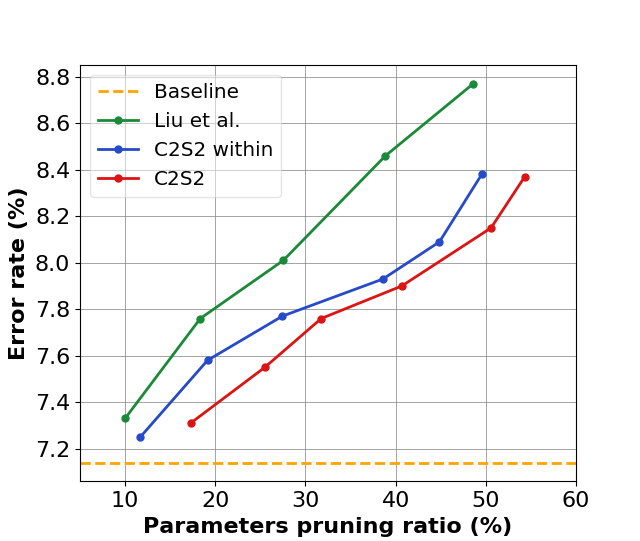}
\caption{Pruning results of ResNet-20 on CIFAR-10. The gap between \cite{LiuLSHYZ17} and C2S2-within tends to become larger as the pruning ratio increases, which indicates the advantage of C2S2 over \cite{LiuLSHYZ17}.} 
\label{fig:resnet20-cifar10}
\end{figure}

As shown in Figure~\ref{fig:resnet20-cifar10}, the pruning result of C2S2 is clearly better than that of \cite{LiuLSHYZ17}. There is a growing trend of the difference between \cite{LiuLSHYZ17} and C2S2-within as the pruning ratio increases, which implies that our channel selection strategy is better than \cite{LiuLSHYZ17}. With $50\%$ parameters pruned, the accuracy drop of our pruned network and that of \cite{LiuLSHYZ17} are $1.0\%$ and $1.6\%$ respectively.

It is mentioned in \cite{Rethink} that fine-tuning a pruned model might lead to worse performance than re-training the pruned model with randomly re-initialized weights. Therefore, we compare the training results using the two strategies: {\em i}) fine-tuning and {\em ii}) re-training with randomly re-initialized weights. As shown in Figure~\ref{fig:retrain_vs_c2s2},
 fine-tuning the pruned model achieves better performance than training from scratch. In other words, our pruning method is not just a kind of architecture search method. It indeed keeps the critical weights that are beneficial to achieving better performance. 

\subsection{Comparison with SSL on CIFAR-10}

\begin{table}[t]
\caption{Pruning ConvNet on CIFAR-10 with C2S2 and SSL.}
\label{table:compare_with_SSL}
\centering
\begin{tabular}{|@{\,}c@{\,}@{\,}c@{\,}@{\,}c@{\,}@{\,}c@{\,}@{\,}c@{\,}|}
\hline
Model & Error & \#channels & FLOPs\,$\downarrow$ & Params\,$\downarrow$ \\
\hline
Baseline & 17.9$\%$ & 32-32-64-10 & - & - \\
\hline
SSL   & 17.9$\%$ & 16-23-63-10 & 69.8$\%$ & 38.4$\%$ \\
\hline
C2S2 & 17.8 $\%$ & 11-16-52-10 & 73.0 $\%$ & 63.8 $\%$ \\
\hline
\end{tabular}
\end{table}

\begin{table}[t]
\caption{Comparison of processing time for pruning ResNet-18 on ImageNet between different methods.We make a sub-training set by randomly sampling data from the original training set, where ``s'' represents the sampling ratio.}
\label{table:avg_time_for_prune}
\centering
\begin{tabular}{|@{\,}c@{\,}|@{\,}c@{\,}|@{\,}c@{\,}|@{\,}c@{\,}|}
\hline
Method & Training dataset & Pruning T & Reduced T $\downarrow$ \\
\hline
\cite{LiuLSHYZ17} & ImageNet & 24hr, 27min & -- \\
\hline
C2S2-within& ImageNet & 23hr, 11min & $0\%$ \\
\hline
C2S2-within& ImageNet ( s=0.5 ) & 13hr, 15min & $42.8\%$ \\
\hline
C2S2-within& ImageNet ( s=0.25 )& 7hr, 10min & $69.1\%$ \\
\hline
\end{tabular}
\end{table}

SSL~\cite{WenWWCL16} is one of the SOTA methods for network pruning which compresses networks by filter-wise, shape-wise and depth-wise pruning. For a fair comparison, we build the ConvNet described in Section 4.2 of SSL~\cite{WenWWCL16} and train it with CIFAR-10 from scratch. The error rate of baseline is 17.85$\%$ which is almost the same as that reported by SSL (17.9$\%$). The pruning results are reported in Table~\ref{table:compare_with_SSL}.
For the structured sparse model obtained by SSL, we report the output channel of each layer by assuming all the redundant convolution filters have been removed.


\subsection{MobileNet on CIFAR-100}

As our method focuses on channel pruning, we implement C2S2 in Torch so that we can accurately compare it with Network Slimming \cite{LiuLSHYZ17}, the current SOTA but available only in Torch implementation. However, we cannot find a publicly available MobileNet pre-trained on ImageNet with Torch implementation. We thus choose to build and train the MobileNet from scratch on CIFAR-100, where its images are rescaled to $224\times 224$ with bicubic interpolation. The redundancy of MobileNet is limited since it is already a compact network. The proposed C2S2 achieves 38.1$\%$ reduction of FLOPs and 22.7$\%$ reduction of parameters with 1$\%$ accuracy loss. In comparison, Network Slimming \cite{LiuLSHYZ17} yields 15.6$\%$ reduction of FLOPs and 15.3$\%$ reduction of parameters with 1$\%$ accuracy loss.


\subsection{ResNet-18 on ImageNet}
We also conduct experiments with ResNet-18 on ImageNet, for further comparison with \cite{LiuLSHYZ17}. For the sake of comparison, we perform within-block compression with \cite{LiuLSHYZ17} and our method. We fine-tune each pruned model for ten epochs with learning rate changing from $10^{-3}$ to $10^{-4}$. The results are plotted in Figure~\ref{fig:resnet18_imagenet}, while more detailed results can be found in the Appendix. 

From Figure ~\ref{fig:resnet18_imagenet}, we can reasonably infer that our method can achieve higher FLOPs reduction than \cite{LiuLSHYZ17} given the same error rate. We further observe that \cite{LiuLSHYZ17} tends to achieve high pruning ratios for high-level layers and low pruning ratios for low-level layers, which leads to less FLOPs reduction. In contrast, our method successfully learns the proper sparsity for each layer and achieves higher FLOPs reduction than \cite{LiuLSHYZ17} given the same error rate. In addition, our pruned model achieves lower error rate than \cite{LiuLSHYZ17} for high parameter reduction, owing to that C2S2 can prune those channels with lower impact on error rate especially for high compression ratios. 
\subsubsection{Runtime Analysis of C2S2}

\begin{figure}[!t]
	\centering
    \includegraphics[width=0.975\linewidth]{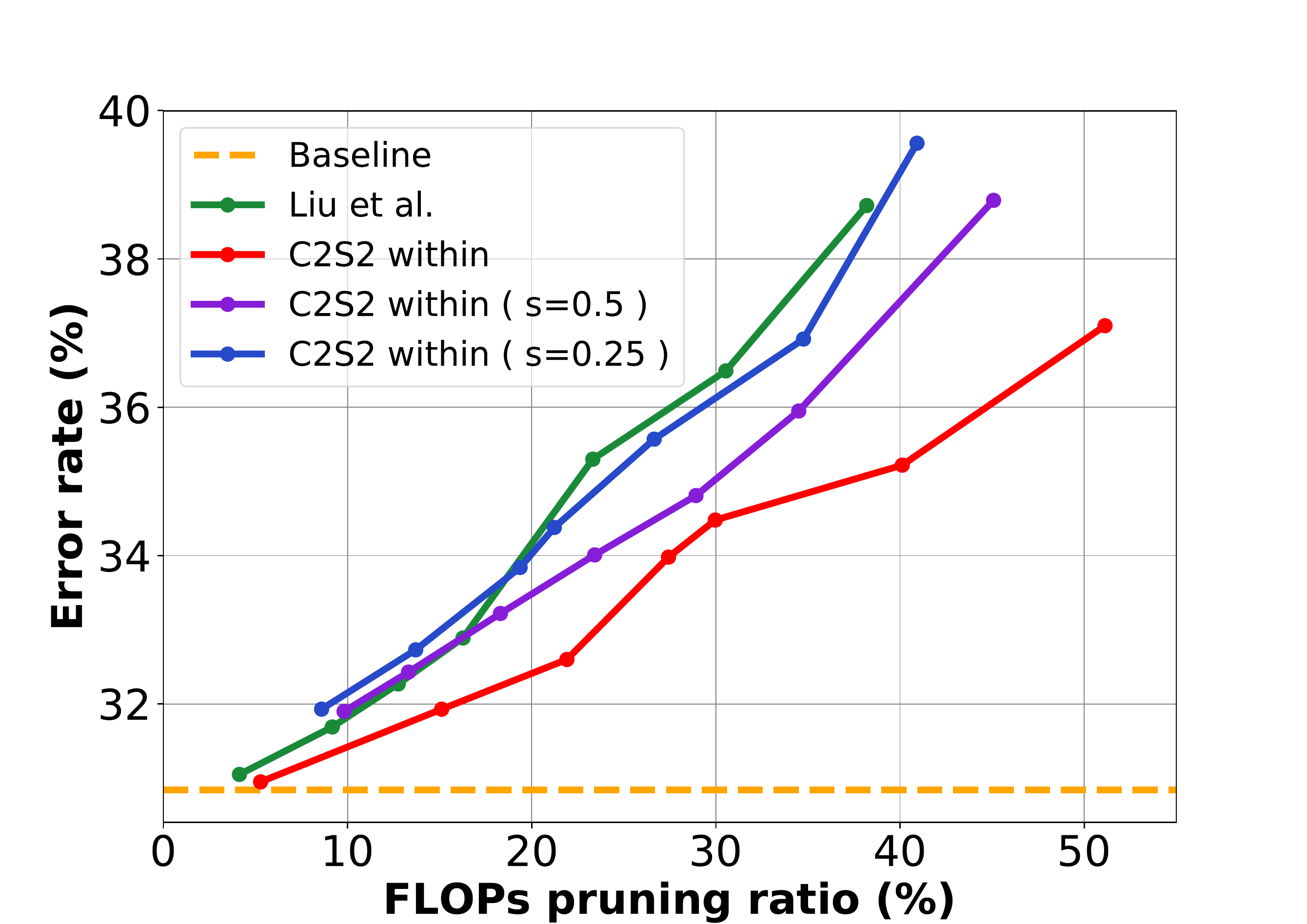}
    \caption{Pruning results of ResNet-18 on ImageNet with complete or partial training data. Where "s" means the sampling ratio.}
    \label{fig:prune_with_subdata}
\end{figure}

In order to compare the efficiency with \cite{LiuLSHYZ17}, we measure the averaged  time taken by C2S2 and \cite{LiuLSHYZ17} to finish the pruning process. The results are reported in Figure \ref{fig:prune_with_subdata} and Table \ref{table:avg_time_for_prune}. The averaged time taken by each algorithm is roughly the same, which makes us to think about how to compress the network faster and still achieve a good accuracy-FLOPs trade-off. We thus perform the pruning process by using a subset of original training set and measure the averaged time for pruning.  We can save 42.8$\%$ time for pruning on averaged, while achieving better accuracy-FLOPs trade-off than \cite{LiuLSHYZ17} by using half of ImageNet training data during the pruning process. We further perform pruning with a quarter of ImageNet training data. The results are comparable to those of \cite{LiuLSHYZ17} but our processing speed is three times faster.


\begin{figure}[!t]
	\centering
    \includegraphics[width=0.485\linewidth]{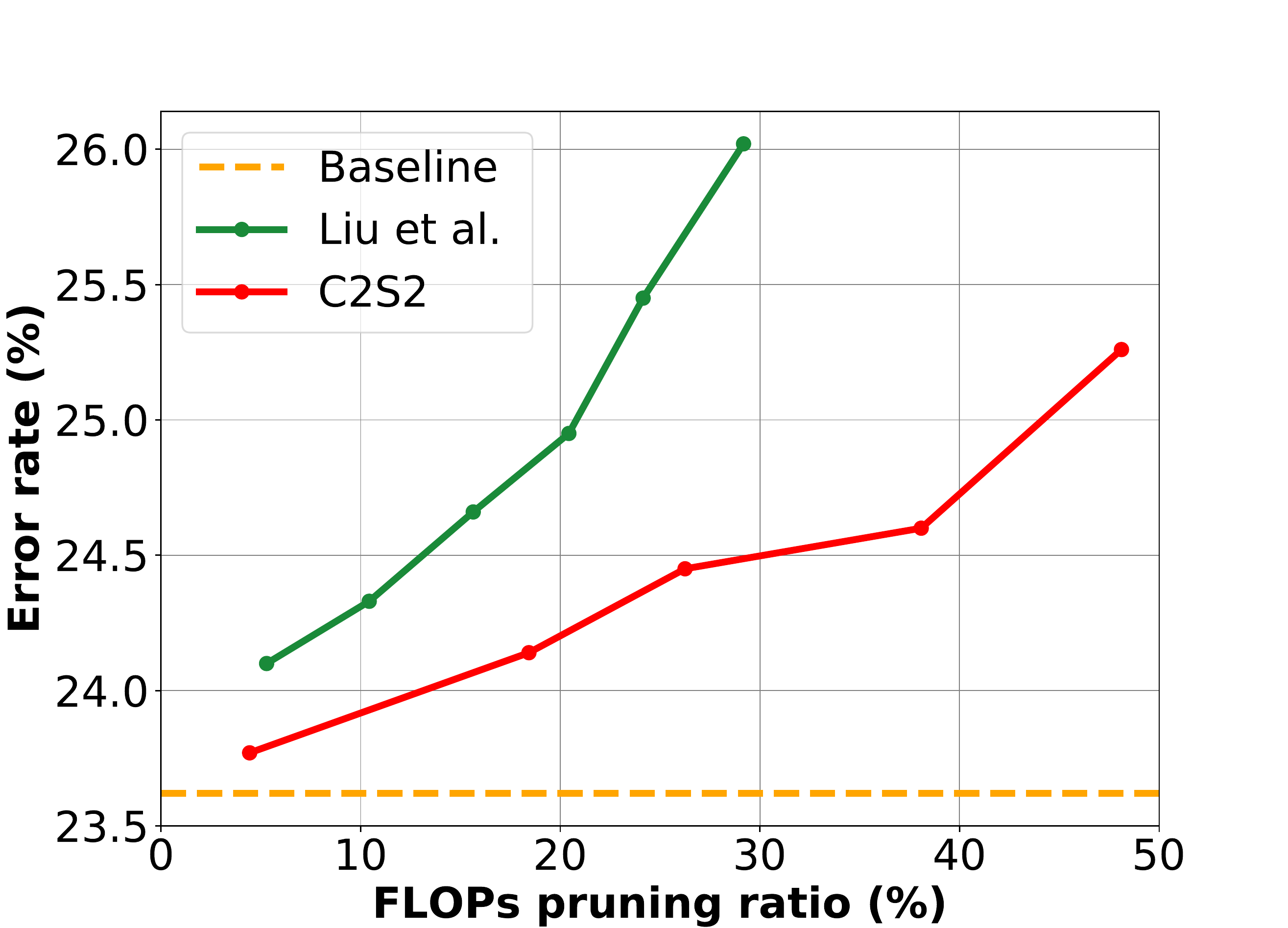}
    \includegraphics[width=0.485\linewidth]{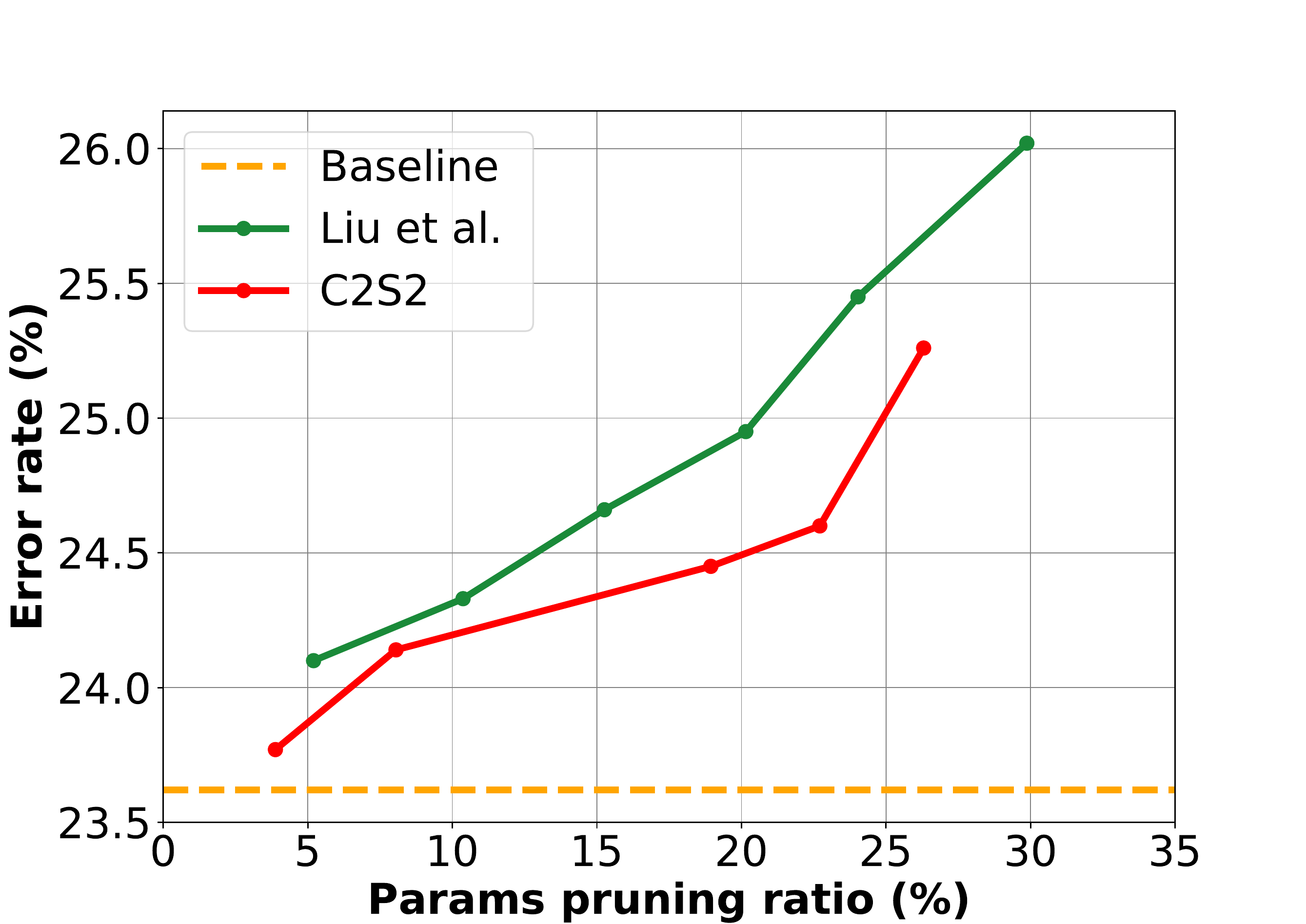}
    \caption{Pruning MobileNet--CIFAR-100 vs. pruning ratios.}
\end{figure}

\begin{figure}[!t]
	\centering
    \includegraphics[width=0.485\linewidth]{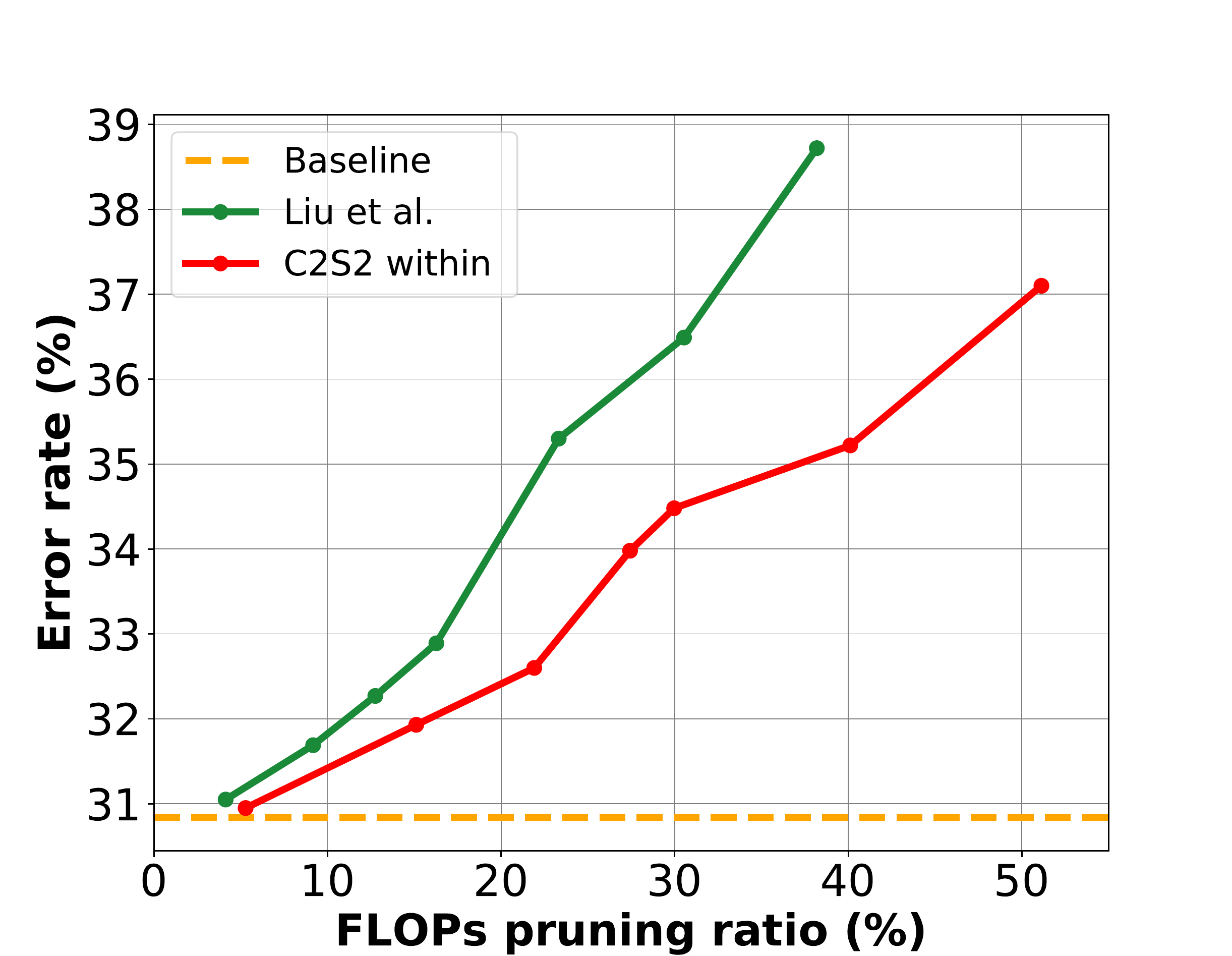}
    \includegraphics[width=0.485\linewidth]{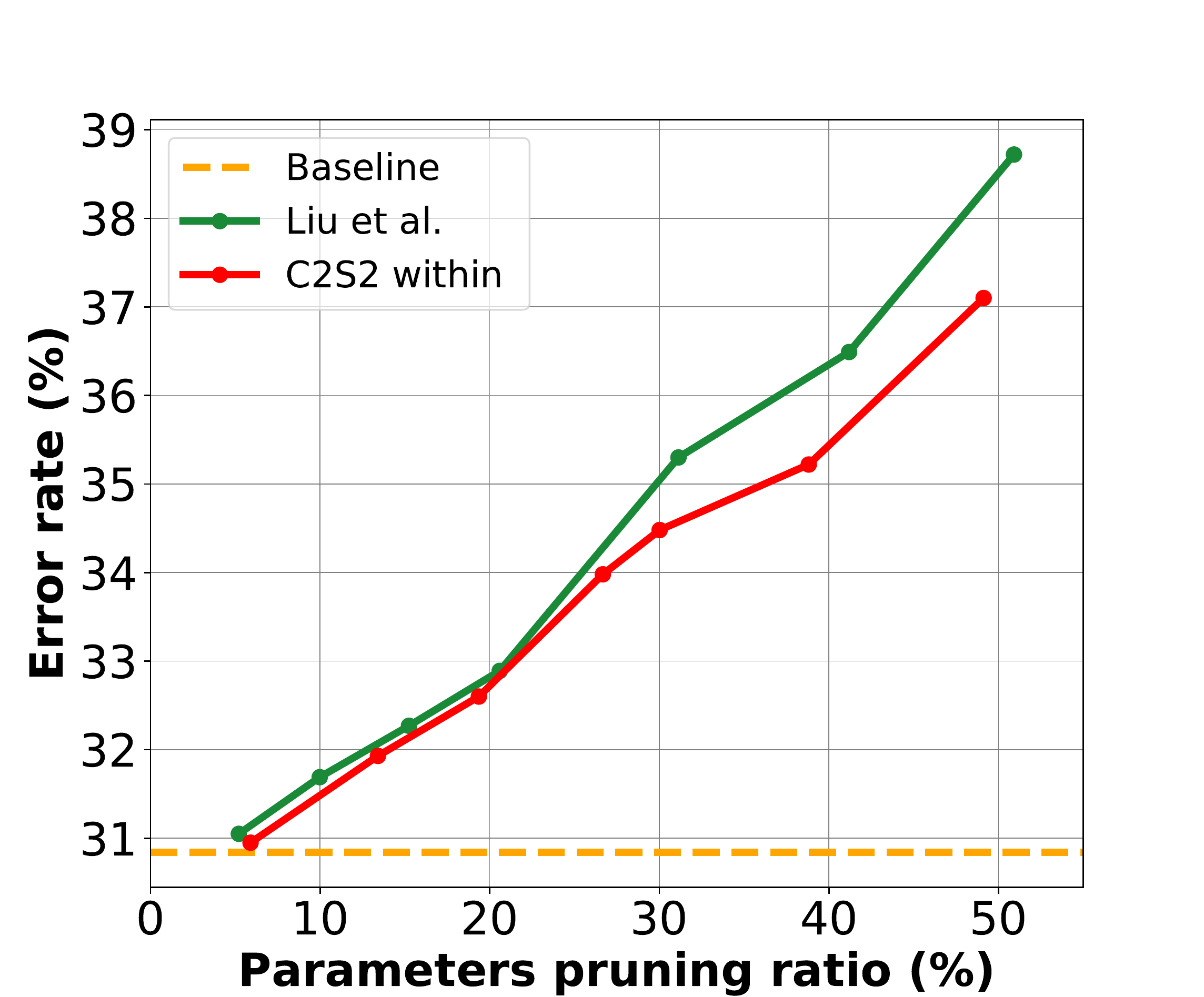}
    \caption{We compress the ResNet-18 on ImageNet with different pruning ratios by changing the value of $c_{r}$.}
    \label{fig:resnet18_imagenet}
\end{figure}

\begin{table*}[t]
\caption{ The pruning results of ResNet-20 on CIFAR-10. The error rate of baseline is 7.14\%. FLOPs\,$\downarrow$ and Params\,$\downarrow$ denote respectively the reduction ratio in the number of FLOPs and the number of parameters.}
\label{table:B1}
\centering
\begin{tabular}{|@{\;}c@{\;}@{\;}c@{\;}@{\;}c@{\;}||@{\;}c@{\;}@{\;}c@{\;}@{\;}c@{\;}||@{\;}c@{\;}@{\;}c@{\;}@{\;}c@{\;}|}
\hline
\multicolumn{3}{|@{}c@{\;}||@{\;}}{C2S2-within} & 
\multicolumn{3}{@{}c@{\;}||@{\;}}{C2S2}& \multicolumn{3}{@{\;}c@{}|}{\cite{LiuLSHYZ17}} \\
\hline
Error rate & FLOPs\,$\downarrow$ & Params\,$\downarrow$ & Error rate & FLOPs\,$\downarrow$ & Params\,$\downarrow$ & Error rate & FLOPs\,$\downarrow$ & Params\,$\downarrow$ \\
\hline
7.25\% & 13.37\% & 11.69\% & 7.31\% & 18.65\% & 17.28\% & 7.33\% & 10.40\% & 10.00\% \\
\hline
7.58\% & 22.22\% & 19.14\% & 7.55\% & 27.25\% & 25.50\% & 7.76\% & 22.22\% & 18.34\% \\
\hline
7.77\% & 34.55\% & 27.43\% & 7.76\% & 38.99\% & 31.75\% & 8.01\% & 31.75\% & 27.54\% \\
\hline
7.93\% & 39.43\% & 38.59\% & 7.90\% & 42.02\% & 40.68\% & 8.46\% & 40.28\% & 38.86\% \\
\hline
8.09\% & 46.97\% & 44.86\% & 8.15\% & 51.64\% & 50.58\% & 8.77\% & 48.41\% & 48.59\% \\
\hline
8.38\% & 52.61\% & 49.54\% & 8.37\% & 54.99\% & 54.31\% & & & \\
\hline
\end{tabular}
\end{table*}

\section{Conclusions}
\label{sec:conclusion}
%

We present a channel selection method to prune redundant parameters of deep networks. Our method automatically identifies the insignificant channels of each layer via a generic learning process. Furthermore, our method can guide itself to learn proper sparsity for each layer with negligible accuracy degradation using the proposed cost-aware mechanism. When the pruning process is completed, we remove the augmented pruning layers and the identified insignificant channels to obtain a pruned model, which is ready for deployment. With C2S2 pruning, we have achieved state-of-the-art reductions in model parameters and FLOPs for all deep-net models evaluated in the experiments. The flexibility of pruning layer enables C2S2 to work with almost all variants of ConvNets, while its dual representations lead to end-to-end training for channel selection. Last but not least, the pruning strategy by C2S2 is driven by maintaining network accuracy, while other approaches accomplish the task by repeatedly exploring pre-specified compressed ratios.

\bibliographystyle{IEEEtran}
\bibliography{C2S2}
\end{document}


\title{C2S2: Cost-aware Channel Sparse Selection for Progressive Network Pruning}
\author{Chih-Yao~Chiu, 
        Hwann-Tzong~Chen,~\IEEEmembership{Member,~IEEE,}
        and~Tyng-Luh~Liu,~\IEEEmembership{Member,~IEEE}}

\appendices

\section{The C2S2 algorithm}
The steps of our method are listed in Algorithm~\ref{alg:C2S2}. 
The following two equations will be referred to by the algorithm. Equation (\ref{eqn:loss_app}) is the objective function for learning pruning weights. Equation (\ref{eqn:mask_app}) illustrates how we update the exponential moving average of training error rate $\cE_{ema}$.
\begin{equation}
	\cL_{all}=\cL (Y,Y_1)+\lambda_{1}\cdot \|P_\ell \|_{1} + \lambda_{2} \cdot \|P_\ell \odot (\bOne - P_\ell)\|_{1}\,,
    \label{eqn:loss_app}
\end{equation}

\begin{equation}
B_\ell(i)=\left\{ \begin{array}{cl} 1\,, & \text{if} \; P_\ell(i)>0.5 \,; \\ 0\,, & \text{otherwise}. \end{array}\right.
\label{eqn:mask_app}
\end{equation}


\section{FCN on PASCAL VOC 2011}

We refer to the architecture design of FCN-8s described in \cite{Long2014} and add batch normalization after each convolution layer.
The FCN-8s baseline is trained on PASCAL VOC 2011 dataset with weight initialized from VGG-16 pre-trained on ImageNet \cite{DengDSLL009}. Both the objective function for training the baseline and the function $\cL(Y,Y_{1})$ in (\ref{eqn:loss_app}) are defined as the sum of pixel-wise cross entropy between each pixel's ground-truth and predicted label over the entire image.
\comment{
\begin{equation}
	\cL_{all}=\cL (Y,Y_1)+\lambda_{1}\cdot \|P_\ell \|_{1} + \lambda_{2} \cdot \|P_\ell \odot (\bOne - P_\ell)\|_{1}\,.
    \label{eqn:loss_app}
\end{equation}
}

\begin{figure}[tbh]
	\centering
	\includegraphics[width=0.95\linewidth]{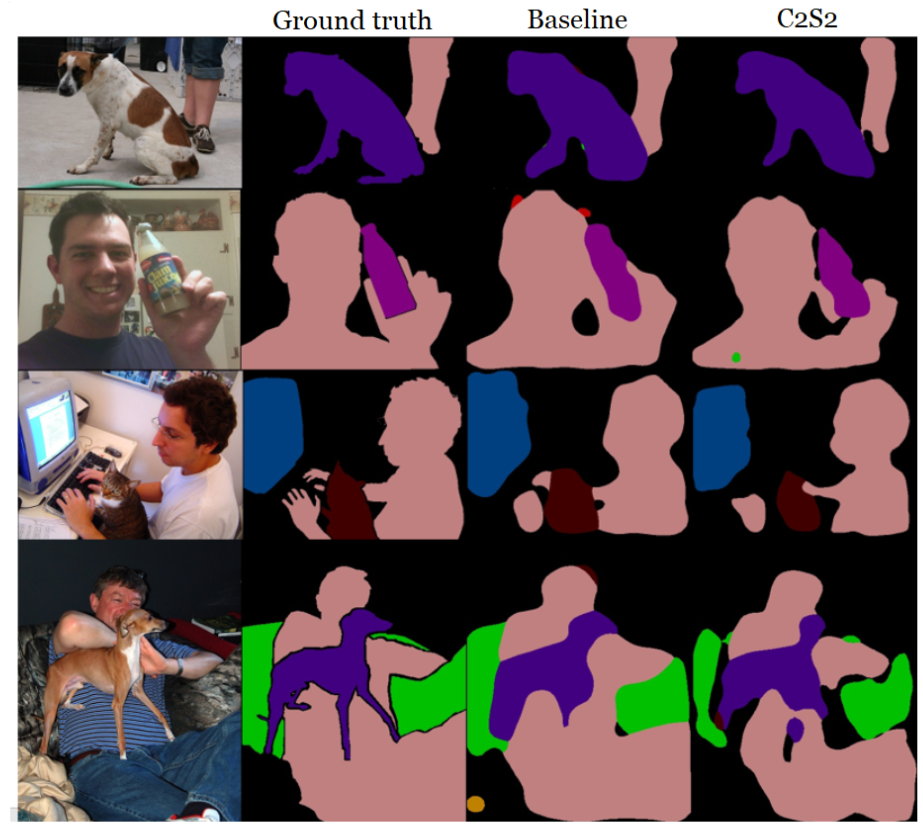}
    \caption{Segmentation results of FCN-8s on PASCAL VOC 2011 validation images. Our pruned model predicts much finer boundaries of objects compared to those of baseline in some cases.}
	\label{figs:seg_results}
\end{figure}

As shown in Table~\ref{table:FCN8s}, we achieve 95.6\% and 63.7\% reduction of model parameters and FLOPs, respectively, without seriously degrading the segmentation performance. 
Figure~\ref{figs:seg_results} illustrates the segmentation results before and after pruning FCN-8s.
The segmentation results of pruned model tend to include ``broken'' objects and reach slightly lower mean IU than the baseline. However, the pruned model can predict finer object boundaries, compared to those of the baseline in several cases like the top two images in Figure~\ref{figs:seg_results}.

\section{Details of pruning results on various network architectures}

\begin{figure}[tbh]
\centering
\includegraphics[width=0.97\linewidth]{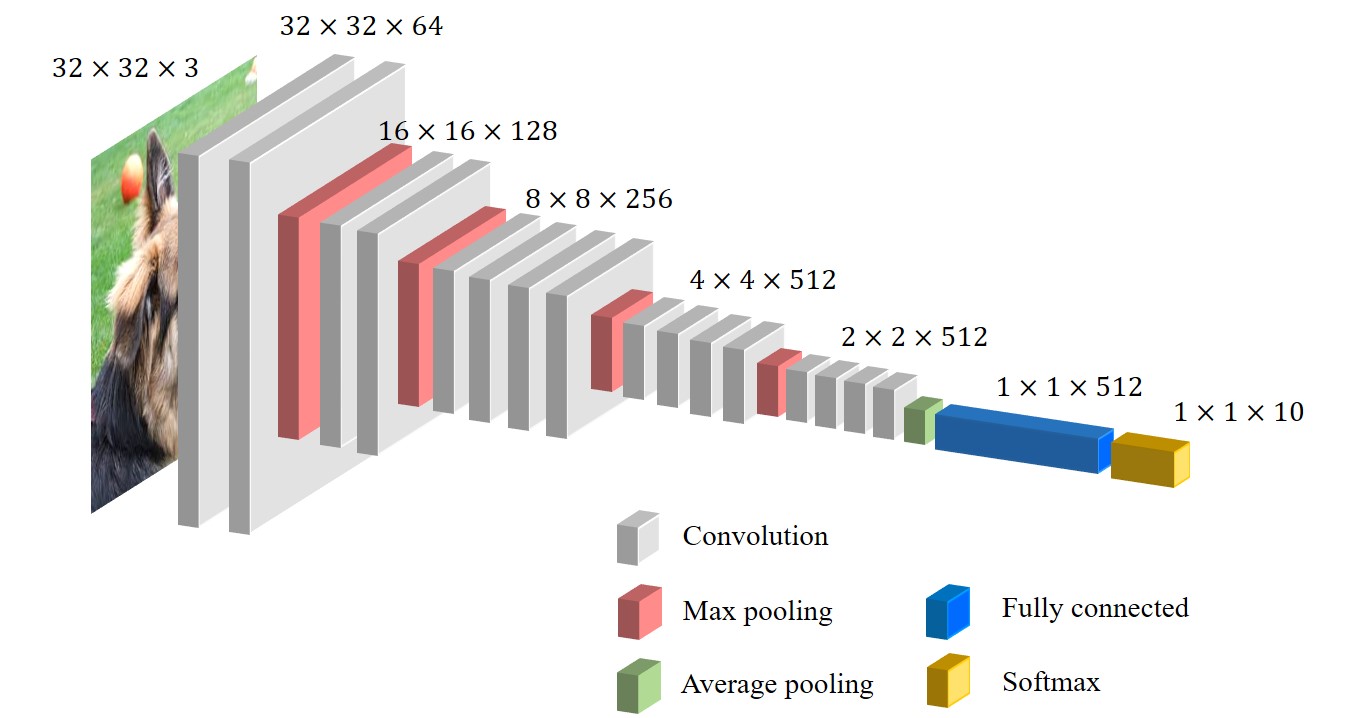}
\caption{The architecture of VGG-Net. It contains 16 convolution layers and 1 fully connected layer. The kernel size of all convolution layers are 3. }
\label{fig:vgg_archi}
\end{figure}

We report the detailed pruning results of VGG-Net, ResNet-20, MobileNet, and ResNet-18 in Tables~\ref{table:A1}-\ref{table:D1}, respectively. We have shown these results as error-rate curves in Figures~5-8 in the main paper. Besides, the architecture of VGG-Net is illustrated in Figure~\ref{fig:vgg_archi}. From the results we can reasonably infer that our method can achieve higher FLOPs and parameters reduction ratios than (Liu et al.2017) given the same error rate.

\begin{algorithm}[t]
  \caption{C2S2: Cost-aware Channel Sparse Selection}
  \label{alg:C2S2}
  \begin{algorithmic}[l]
    \REQUIRE Dataset $D=(\bX, \bY)$ and ConvNet $M$
    \ENSURE A compressed model of $M$
    \STATE Compute mean training error rate of $M$ over $D$ as $\cE_{base}$
	\STATE Construct augmented network $\widetilde{M}$ with $L$ pruning layers
    \STATE Initialize $P_\ell \leftarrow \cN(\mu=1, \sigma^2=0.01)$ ,  $B_\ell \leftarrow \bOne$, \; $\forall \ell$
    \STATE Initialize step , updateFrequency , $\alpha , \cE_{ema}$
    \STATE //Start to prune the network
    \FOR{$\ell = 1, \dots, L$}
        \STATE State $\leftarrow$ {\tt Pruning} 
        \WHILE{{\tt True}}
            \STATE $X$, $Y$ = getTrainData($D$)
    		\IF {mod(step, updateFrequency) == 0}
                \STATE // Update the pruning layer
    			\STATE $Y_1 = \widetilde{M} (X, \mathbf{W},\mathbf{P})$ 
                \STATE Compute $\cL_{all}$ from (\ref{eqn:loss_app})
                \STATE $\{\widetilde{M}, P_\ell\}$ $\leftarrow$ Optimize($\cL_{all}$, $P_\ell$, $\widetilde{M}$) 
                \STATE $B_\ell$ = Binarize($P_\ell$, $0.5$) from (\ref{eqn:mask_app})
            \ENDIF
            \STATE
            \STATE // Update all convolution layers
            \STATE $Y_2 = \widetilde{M}(X,\bW,\bB$) 
            \STATE $\mathbf{W}$ = Optimize($\cL(Y, Y_2)$), $\mathbf{W}$)
            \STATE 
            \STATE // Cost-aware mechanism
            \STATE // $\cE(Y,Y_2)$ would return the batch error rate.
            \STATE $\cE_{ema}\leftarrow(1-\alpha)\cE_{ema} + \alpha\cE(Y, Y_2)\,\,$
            \IF {State == {\tt Pruning}}
                \IF {$\cE_{ema}>c_p\times \cE_{base}$}
                   \STATE $\lambda_{1}\leftarrow -\lambda_{1}$, State $\leftarrow$ {\tt Restoring}
               \ENDIF
            \ELSE
            \STATE // State == Restoring
            \IF {$\cE_{ema} < c_r\times \cE_{base}$}
		       \STATE {break} //Go to prune next layer 
            \ENDIF
          \ENDIF
        \STATE step $\leftarrow$ step + 1
        \ENDWHILE
    \ENDFOR
    \STATE Prune channels according to $\mathbf{B}$
    \STATE Fine-tune the compressed model
  \end{algorithmic}
\end{algorithm}

\begin{table}[tbh]
\caption{The pruning results of FCN-8s on PASCAL VOC 2011. Since the shape of input image for FCN-8s is not fixed, we report the average FLOPs over 1{,}111 validation images. The setting of hyper-parameters is $c_{p} = 4.0 ,c_{r}=2.0$. }
\label{table:FCN8s}
\centering
\begin{tabular}{|c|c|c|c|c|}
\hline
Model & Pixel acc. & Mean acc. & Mean IU & F.w. IU \\
\hline
Baseline & 90.4\% & 74.5\% & 57.5\% & 82.1\% \\
C2S2 & 88.9\% & 73.2\% & 54.8\% & 81.2\% \\
\hline
\hline
Model & \#Params & $\downarrow$ Ratio & FLOPs & $\downarrow$ Ratio \\
\hline
Baseline & 1.34E+08 & - & 3.31E+11 & - \\
C2S2 & 5.86E+06 & 95.6\% & 1.20E+11 & 63.7\% \\
\hline
\end{tabular}
\end{table}

\begin{table}[tbh]
\caption{ Pruning results of VGG-Net on CIFAR-10. The error rate of baseline is 6.20\%. FLOPs\,$\downarrow$ and Params\,$\downarrow$ denote respectively the reduction ratio in the number of FLOPs and the number of parameters.}
\label{table:A1}
\centering
\begin{tabular}{|@{\,}c@{\,}@{\,}c@{\,}@{\,}c@{\,}||@{\,}c@{\,}@{\,}c@{\,}@{\,}c@{\,}|}
\hline
\multicolumn{3}{|@{}c@{\,}||@{\,}}{C2S2} & \multicolumn{3}{@{\,}c@{}|}{\cite{LiuLSHYZ17}} \\
\hline
Error rate & FLOPs\,$\downarrow$ & Params\,$\downarrow$ & Error rate & FLOPs\,$\downarrow$ & Params\,$\downarrow$ \\
\hline
6.03\% & 53.66\% & 89.68\% & 6.08\% & 37.69\% & 67.48\% \\
\hline
6.10\% & 58.70\% & 91.00\% & 6.06\% & 44.24\% & 77.48\% \\
\hline
6.22\% & 60.95\% & 92.11\% & 6.05\% & 51.72\% & 88.79\% \\
\hline
6.68\% & 66.00\% & 93.40\% & 6.19\% & 55.55\% & 90.73\% \\
\hline
6.79\% & 69.06\% & 93.95\% & 6.25\% & 57.79\% & 91.59\% \\
\hline
7.13\% & 72.04\% & 94.41\% &        &         & \\
\hline
\end{tabular}
\end{table}

\begin{table}[ht]
\caption{ Pruning results of ResNet-20 on CIFAR-10. The error rate of baseline is 7.14\%. FLOPs\,$\downarrow$ and Params\,$\downarrow$ denote respectively the reduction ratio in the number of FLOPs and the number of parameters.}
\label{table:B1}
\centering
\begin{tabular}{|@{\;}c@{\;}@{\;}c@{\;}@{\;}c@{\;}||@{\;}c@{\;}@{\;}c@{\;}@{\;}c@{\;}|}
\hline
\multicolumn{3}{@{}c@{\;}||@{\;}}{C2S2}& \multicolumn{3}{@{\;}c@{}|}{\cite{LiuLSHYZ17}} \\
\hline
 Error rate & FLOPs\,$\downarrow$ & Params\,$\downarrow$ & Error rate & FLOPs\,$\downarrow$ & Params\,$\downarrow$ \\
\hline
 7.31\% & 18.65\% & 17.28\% & 7.33\% & 10.40\% & 10.00\% \\
\hline
 7.55\% & 27.25\% & 25.50\% & 7.76\% & 22.22\% & 18.34\% \\
\hline
 7.76\% & 38.99\% & 31.75\% & 8.01\% & 31.75\% & 27.54\% \\
\hline
 7.90\% & 42.02\% & 40.68\% & 8.46\% & 40.28\% & 38.86\% \\
\hline
 8.15\% & 51.64\% & 50.58\% & 8.77\% & 48.41\% & 48.59\% \\
\hline
 8.37\% & 54.99\% & 54.31\% & & & \\
\hline
\end{tabular}
\comment{
\begin{tabular}{|@{\;}c@{\;}@{\;}c@{\;}@{\;}c@{\;}||@{\;}c@{\;}@{\;}c@{\;}@{\;}c@{\;}||@{\;}c@{\;}@{\;}c@{\;}@{\;}c@{\;}|}
\hline
\multicolumn{3}{|@{}c@{\;}||@{\;}}{C2S2-within} & 
\multicolumn{3}{@{}c@{\;}||@{\;}}{C2S2}& \multicolumn{3}{@{\;}c@{}|}{\cite{LiuLSHYZ17}} \\
\hline
Error rate & FLOPs\,$\downarrow$ & Params\,$\downarrow$ & Error rate & FLOPs\,$\downarrow$ & Params\,$\downarrow$ & Error rate & FLOPs\,$\downarrow$ & Params\,$\downarrow$ \\
\hline
7.25\% & 13.37\% & 11.69\% & 7.31\% & 18.65\% & 17.28\% & 7.33\% & 10.40\% & 10.00\% \\
\hline
7.58\% & 22.22\% & 19.14\% & 7.55\% & 27.25\% & 25.50\% & 7.76\% & 22.22\% & 18.34\% \\
\hline
7.77\% & 34.55\% & 27.43\% & 7.76\% & 38.99\% & 31.75\% & 8.01\% & 31.75\% & 27.54\% \\
\hline
7.93\% & 39.43\% & 38.59\% & 7.90\% & 42.02\% & 40.68\% & 8.46\% & 40.28\% & 38.86\% \\
\hline
8.09\% & 46.97\% & 44.86\% & 8.15\% & 51.64\% & 50.58\% & 8.77\% & 48.41\% & 48.59\% \\
\hline
8.38\% & 52.61\% & 49.54\% & 8.37\% & 54.99\% & 54.31\% & & & \\
\hline
\end{tabular}
}
\end{table}


\begin{table}[h]
\caption{ Pruning results of MobileNet on CIFAR-100. The error rate of baseline is 23.62\%. FLOPs\,$\downarrow$ and Params\,$\downarrow$ denote respectively the reduction ratio in the number of FLOPs and the number of parameters.}
\label{table:C1}
\centering
\begin{tabular}{|@{\;}c@{\;}@{\;}c@{\;}@{\;}c@{\;}||@{\;}c@{\;}@{\;}c@{\;}@{\;}c@{\;}|}
\hline
\multicolumn{3}{|@{}c@{\;}||@{\;}}{C2S2} & \multicolumn{3}{@{\;}c@{}|}{\cite{LiuLSHYZ17}} \\
\hline
Error rate & FLOPs\,$\downarrow$ & Params\,$\downarrow$ & Error rate & FLOPs\,$\downarrow$ & Params\,$\downarrow$ \\
\hline
23.77\% & 4.44\% & 3.88\% & 24.10\% & 5.29\% & 5.20\% \\
\hline
24.14\% & 18.43\% & 8.05\% & 24.33\% & 10.43\% & 10.37\% \\
\hline
24.45\% & 26.25\% & 18.94\% & 24.66\% & 15.64\% & 15.26\% \\
\hline
24.60\% & 38.08\% & 22.71\% & 24.95\% & 20.43\% & 20.15\% \\
\hline
25.26\% & 48.11\% & 26.30\% & 25.45\% & 24.15\% & 24.03\% \\
\hline
 & & & 26.02\% & 29.18\% & 29.87\% \\
\hline
\end{tabular}
\end{table}

\begin{table}[h]
\caption{ Pruning results of ResNet-18 on ImageNet. The error rate of baseline is 30.84\%. FLOPs\,$\downarrow$ and Params\,$\downarrow$ denote respectively the reduction ratio in the number of FLOPs and the number of parameters.}
\label{table:D1}
\centering
\begin{tabular}{|@{\;}c@{\;}@{\;}c@{\;}@{\;}c@{\;}||@{\;}c@{\;}@{\;}c@{\;}@{\;}c@{\;}|}
\hline
\multicolumn{3}{|@{}c@{\;}||@{\;}}{C2S2-within} & \multicolumn{3}{@{\;}c@{}|}{\cite{LiuLSHYZ17}} \\
\hline
Error rate & FLOPs\,$\downarrow$ & Params\,$\downarrow$ & Error rate & FLOPs\,$\downarrow$ & Params\,$\downarrow$ \\
\hline
30.95\% & 5.28\% & 5.90\% & 31.05\% & 4.13\% & 5.12\% \\
\hline
31.93\% & 15.11\% & 13.41\% & 31.69\% & 9.17\% & 9.98\% \\
\hline
32.60\% & 21.91\% & 19.36\% & 32.27\% & 12.75\% & 15.24\% \\
\hline
33.98\% & 27.43\% & 26.67\% & 32.89\% & 16.27\% & 20.59\% \\
\hline
34.48\% & 29.97\% & 30.02\% & 35.30\% & 23.32\% & 31.13\% \\
\hline
35.22\% & 40.12\% & 38.81\% & 36.49\% & 30.54\% & 41.19\% \\
\hline
37.10\% & 51.13\% & 49.12\% & 38.72\% & 38.19\% & 50.91\% \\
\hline
\end{tabular}
\end{table}

\bibliographystyle{IEEEtran}
\bibliography{C2S2}